# COMPUTER VISION BASED GROUP ACTIVITY DETECTION AND ACTION SPOTTING

## PROJECT REPORT – GROUP 17

**GROUP MEMBERS**

**E/18/334 SIVALINGAM N**

**E/18/350 THAMAYANTHI M**

**E/18/415 SIVASTHIGAN S**

**SUPERVISORS:**

**PROF. G.M.R.I. GODALIYADDA**

**PROF. M.P.B. EKANAYAKE**

**PROF. H.M.V.R. HERATH**

**Department of Electrical and Electronic Engineering**

**Faculty of Engineering**

**University of Peradeniya**

**Sri Lanka**

**MARCH 2024**

# ACKNOWLEDGEMENT

The successful completion of this semester's project would not have been possible without the generous support and guidance of numerous individuals, for which we are deeply grateful. We extend our heartfelt appreciation to our esteemed project supervisors, namely:

- Prof. G.M.R.I Godaliyadda
- Prof. H.M.V.R Herath
- Prof. M.P.B. Ekanayake

Their exceptional mentoring, constant encouragement, and valuable insights were instrumental in helping us accomplish our undergraduate project within the stipulated time frame.

Additionally, we would like to extend our sincere thanks to the Coordinator of the undergraduate project, Dr. Ruwan Ranaweera, for his meticulous evaluation of our work and unwavering support.

Furthermore, we express our gratitude to all the lecturers, colleagues, and other individuals who contributed their creative ideas, encouragement, and assistance, playing a significant role in the success of our semester's project.

With profound appreciation,

Narthana Sivalingam (E/18/334)
Thamayanthi Mahendranathan (E/18/350)
Santhirarajah Sivasthigan (E/18/415)

Department of Electrical and Electronic Engineering,
Faculty of Engineering,
University of Peradeniya,
Peradeniya 20400,
Sri Lanka.

# CONTENTS

# List of Figures

# List of Tables

# CHAPTER 1: PROBLEM STATEMENT

In recent years, the rapid advancement of video surveillance technology and the ubiquity of video data have heightened the need for intelligent systems capable of analyzing and interpreting complex group activities and interactions among individuals within diverse scenarios. The ability to detect and recognize group activities accurately is vital across a range of applications, from public safety and security to sports analysis, urban planning, and beyond. Traditional methods, reliant on manual analysis or simplistic algorithmic approaches, fall short in addressing the dynamic and nuanced nature of group interactions, leading to inefficiencies, inaccuracies, and missed critical insights.

The project titled "Computer Vision-Based Group Activity Detection and Action Spotting" aims to tackle these challenges by leveraging cutting-edge deep learning technologies to automate the process of detecting and recognizing group activities within video streams. This problem statement outlines the scope, objectives, and methodologies of the project, underscoring its significance in the context of contemporary technological needs.

The core objective of this project is to develop a robust system that can automatically detect, recognize, and analyze group activities by interpreting complex interactions among individuals in video data. To achieve this, the project will employ Convolutional Neural Networks (CNNs)—specifically, MobileNet, Inception V3, and VGG16—for feature extraction, which is a critical step in understanding the visual content of videos. These CNN models are chosen for their proven effectiveness in capturing high-level visual features across a wide range of contexts.

A novel aspect of our approach is the use of Mask R-CNN for two key objectives. Firstly, Mask R-CNN is employed to extract mask data for the actors in the images, allowing for precise identification of individuals within the video frames. Secondly, it is used to determine the exact boundary box coordinates for these actors, ensuring accurate localization within the scene. The boundary box coordinates extracted from Mask R-CNN are then used to filter the feature map of actors in the scene, derived from the entire frame's feature map extracted through CNN.

Further, the mask data extracted by Mask R-CNN is convolved with the boundary box feature map of each actor, creating a masked feature map for each actor across a sequence of images. This process refines the representation of each individual, enhancing the system's ability to recognize and analyze their interactions within the group.

Additionally, the project will create an Actor Relation Graph (ARG) using these masked feature maps, representing the dynamic interactions among individuals within a scene and encoding both spatial and temporal relationships. On this ARG, Graph Convolutional Networks (GCNs) will be applied to predict group activities, as GCNs are particularly adept at handling graph-structured data, making them ideal for analyzing the complex, interconnected nature of group interactions.

To further enhance the accuracy of our system, we will incorporate appearance similarity search techniques, including Normalized Cross-Correlation (NCC), Sum of Absolute Differences (SAD), and dot product measures. These methods will be used to identify and track individuals across frames, facilitating the recognition of group activities by analyzing the consistency and changes in appearance among actors within the scene.

This project is poised to make significant contributions to the field of computer vision by addressing a critical gap in current video analysis technologies—the ability to automatically and accurately detect and interpret complex group activities. By combining advanced CNN models for feature extraction, precise localization techniques with Mask R-CNN, innovative graph-based analysis, and sophisticated similarity search algorithms, we aim to develop a system that not only enhances the efficiency and accuracy of group activity detection but also opens new avenues for research and application in various domains.

In conclusion, "Computer Vision-Based Group Activity Detection and Action Spotting" represents a comprehensive and innovative approach to understanding the rich tapestry of human interactions captured in video data. Through the implementation and evaluation of this system, we aim to push the boundaries of what is possible in automated video surveillance and analysis, contributing valuable insights and tools to the field.

# CHAPTER 2: LITERATURE REVIEW

## 2.1 Object Detection

Object detection, a pivotal aspect of computer vision, involves identifying and locating objects within digital images. This field has seen rapid advancements, primarily driven by deep learning algorithms, which can be categorised into two principal groups based on their operational methodology.

1. Algorithms that employ a two-stage approach, where the initial step involves identifying potential regions of interest (RoIs) within an image, followed by the classification of these regions using Convolutional Neural Networks (CNNs)[1]. This method, although methodical, tends to be slower due to the necessity of processing multiple regions individually. Prominent examples of this category include the Region-based Convolutional Neural Network (R-CNN)[2] and its evolutionary successors: Fast R-CNN[3], which introduces a more efficient way to share computations; Faster R-CNN[4], which further improves speed by integrating region proposal within the network; and Mask R-CNN[5], which extends Faster R-CNN by adding a branch for segmenting objects at a pixel level, making it suitable for tasks requiring precise object localization.

2. Algorithms that adopt a regression approach to directly predict both the classes and the bounding boxes of objects across the entire image in a single pass, significantly enhancing the speed and efficiency of the detection process. The most notable example of this approach is YOLO (You Only Look Once), which revolutionised the field by drastically reducing the computation time without compromising the accuracy significantly. Subsequent iterations, such as YOLOv2 and YOLOv3, have continued to refine this balance, offering improved detection precision and speed.

These contrasting approaches highlight the trade-offs between accuracy and efficiency in object detection algorithms. While two-stage methods are generally more accurate, single-pass detectors offer the advantage of speed, making them more suitable for applications requiring real-time processing. As the field progresses, ongoing research and development aim to further bridge these gaps, promising even more robust and efficient object detection techniques.

### 2.1.1 YOLO- You Only Look Once

The YOLO (You Only Look Once) [6] framework introduces a novel approach to object detection by reframing the problem as a single regression task that directly maps image pixels to bounding box coordinates and class probabilities. This method stands out for its simplicity and speed, leveraging a single convolutional neural network (CNN) to predict multiple bounding boxes and their corresponding class probabilities simultaneously [7].

**Here's a structured overview of YOLO's key concepts and mechanisms:**

1. Unified Detection: YOLO unifies the traditionally separate tasks of object localization and classification by using a single neural network to do both simultaneously. This allows the model to view the entire image during both training and testing phases, enabling it to learn contextual information about object classes and their appearance within the full image scope [8].

2. Grid Division: The input image is divided into an SxS grid. Each grid cell is responsible for predicting the objects whose centers fall within it. This spatial division means that the detection is localized, with each part of the image being examined for relevant object features [9].

3. Bounding Box Predictions: Within each grid cell, YOLO predicts multiple bounding boxes and assigns confidence scores to each. The confidence score reflects the model's certainty that a box contains an object and the accuracy of the box itself. Mathematically, the confidence score is defined as Pr(Object) * IoU (Intersection over Union), where Pr(Object) indicates the probability of an object's presence, and IoU measures the accuracy of the box[10].

4. Box Dimension Predictions: Each predicted bounding box is described by five attributes which are center coordinates (x, y), width (w), height (h), and a confidence score. The center coordinates are relative to the grid cell, while the width and height are relative to the entire image. This prediction strategy enables YOLO to localize objects precisely within the image [11].

5. Class Probability Predictions: Alongside bounding boxes, each grid cell predicts conditional class probabilities, PR(Class_i | Object), for the presence of each object class. These probabilities are conditioned on the assumption that an object is present in

the grid cell. YOLO simplifies this to predict only one set of class probabilities per grid cell, regardless of the number of bounding boxes B [12].

6. Performance and Efficiency: YOLO's design allows it to process images extremely quickly, making it suitable for real-time applications. Unlike methods that rely on a sliding window or region proposals, YOLO's global view of the image reduces the number of background errors—situations where the model incorrectly identifies background patches as objects [13].

7. Error Reduction: Compared to methods like Fast R-CNN, YOLO makes fewer background errors by leveraging its comprehensive view of the image context. This integrated approach helps the model distinguish between relevant objects and background noise more effectively [14].

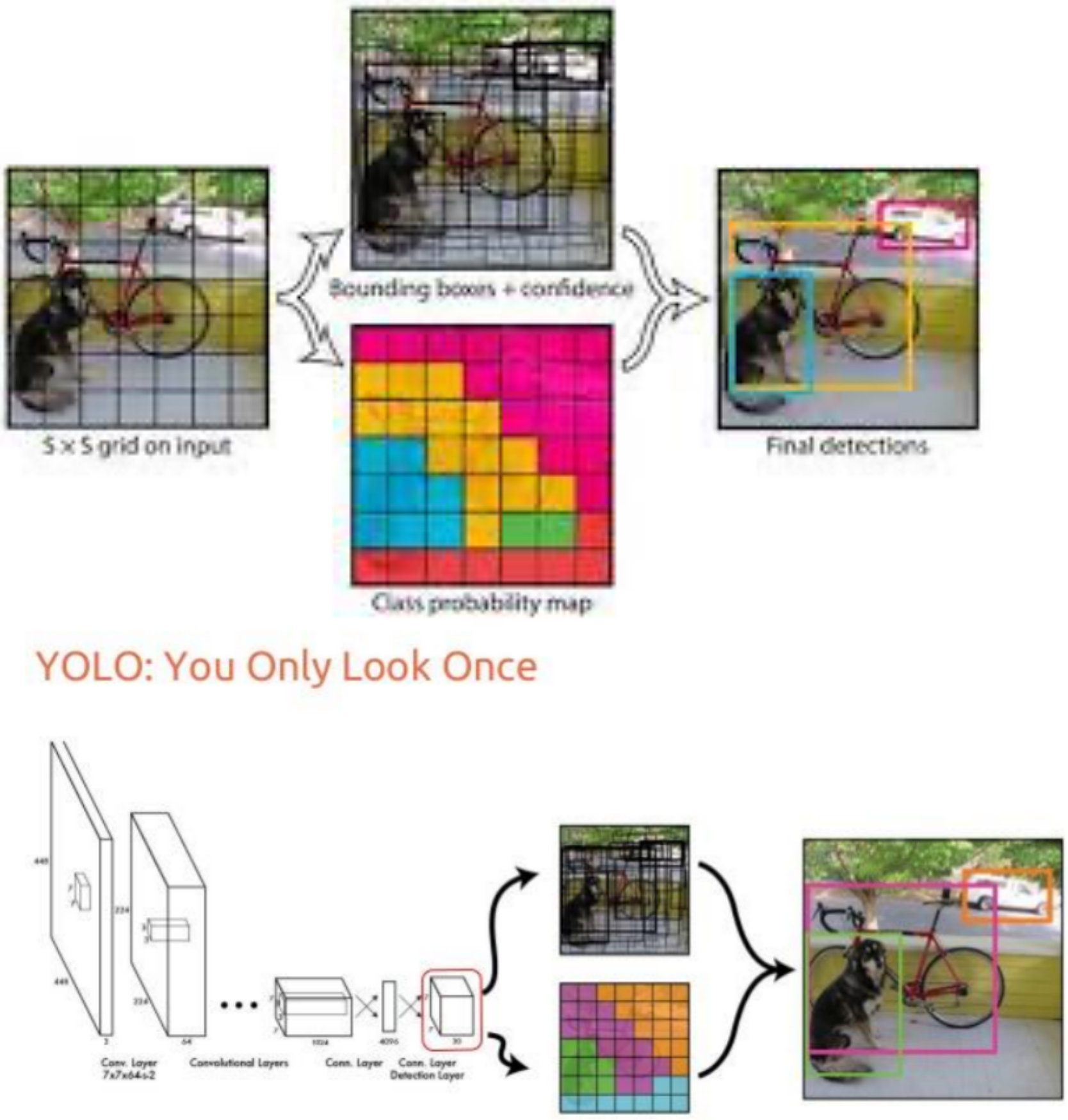

*Figure 2. 1: YOLO Detection Procedure*

YOLO's innovative approach to object detection through a single convolutional network simplifies the detection process, enhances speed, and maintains a high level of accuracy as shown in Figure 01. Its ability to integrate contextual information and directly predict bounding box parameters and class probabilities in one evaluation makes it a robust solution for real-time object detection challenges.

### 2.1.2 CNN Network + ROI Align

Convolutional Neural Networks (CNNs) combined with Region of Interest (RoI) Align for object detection explores the integration of these advanced techniques in computer vision. This review delves into how these methodologies enhance the accuracy and efficiency of object detection systems.

Convolutional Neural Networks have revolutionised the field of image recognition and classification due to their ability to learn hierarchical feature representations from images. The initial layers capture basic features like edges, while deeper layers identify complex patterns. The introduction of CNNs to object detection tasks marked a significant improvement in detection rates, reducing the reliance on hand-crafted features.

Region of Interest techniques are pivotal in object detection, focusing the detection model's attention on specific parts of the image that are likely to contain objects. This selective attention mechanism improves both the speed and accuracy of object detection by reducing the computational load and minimising background noise. RoI Align [15], an improvement over the earlier RoI Pooling, addresses the misalignment issue between the RoI and the extracted features. By using bilinear interpolation, RoI Align ensures that the extracted features are precisely aligned with the original RoI, leading to more accurate object detection.

The combination of CNNs with RoI Align has set new benchmarks in object detection tasks. This integration allows for precise localization and classification of objects within an image, even in challenging conditions such as occlusions, varying scales, and complex backgrounds.

## 2.2 Convolutional Neural Network

Convolutional Neural Networks (CNNs) are a class of deep neural networks, most applied to analysing visual imagery. They have been highly successful in various tasks in computer vision, including image and video recognition, image classification, medical image analysis, and more. The architecture of the CNN is given in Figure 02.

Key Features:

1. Layered Architecture: CNNs are composed of an input layer, multiple hidden layers, and an output layer. The hidden layers include convolutional layers, activation functions, pooling layers, fully connected layers, and normalization layers.

2. Convolutional Layers: Utilize filters or kernels to perform convolution operations, capturing spatial features such as edges, textures, and shapes.
3. Pooling Layers: Reduce the spatial size of the representation, decreasing the number of parameters and computation in the network, thus improving efficiency.
4. Activation Functions: Introduce non-linearity into the network, allowing it to learn complex patterns. Commonly used functions include ReLU (Rectified Linear Unit).

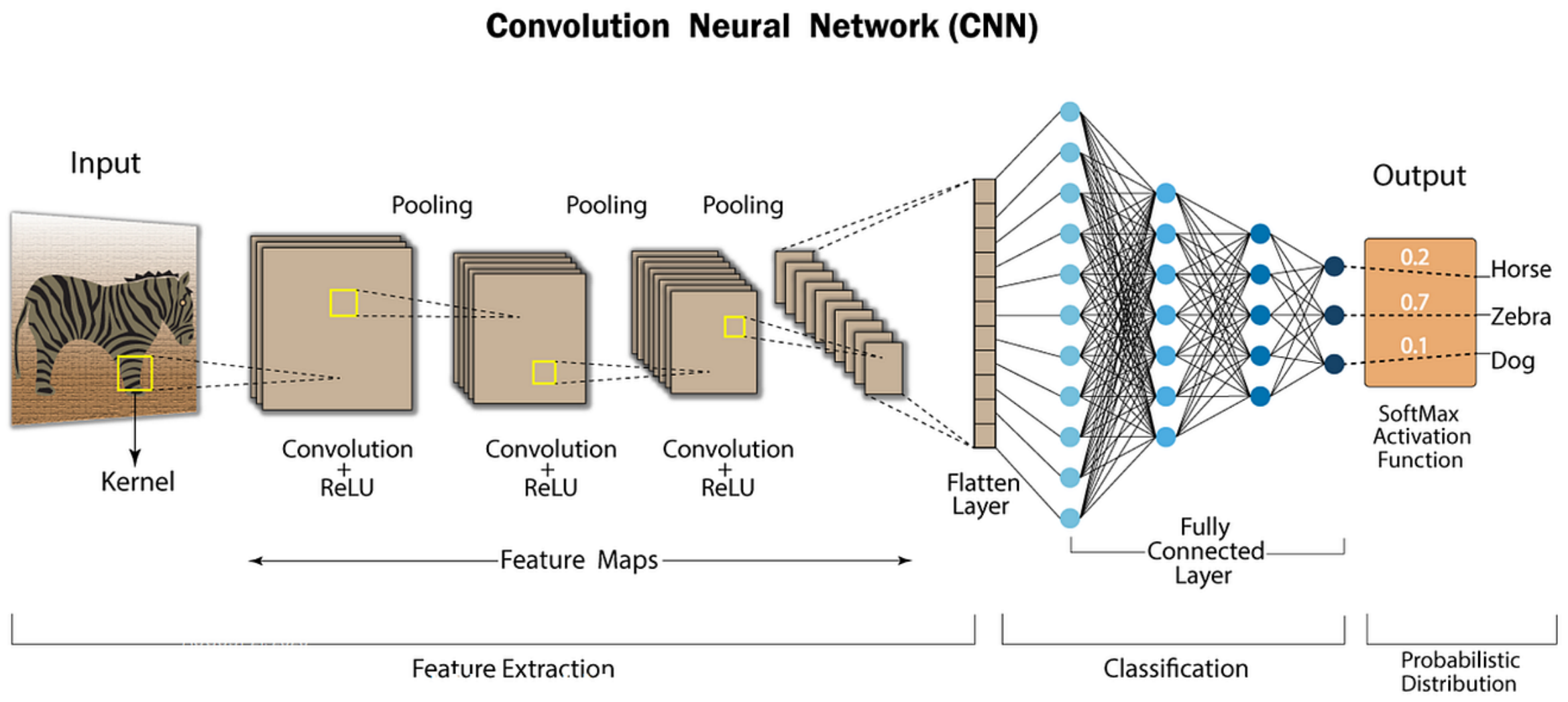

*Figure 2. 2: Architecture of CNN*

Convolution Operation,

$$F(i,j) = (G * H)(i,j) = \sum_m \sum_n G(m,n)H(i-m, j-n)$$

where

F - output feature map
G - input image
H - kernel.

Pooling: Commonly max pooling, where the maximum value over a spatial neighbourhood is taken to represent the region.

## 2.2.1 MobileNet

MobileNet is designed for mobile and embedded vision applications. It is efficient in terms of computation and memory, utilizing depth-wise separable convolutions to reduce the number of parameters and computational cost.

The architecture of mobilenet, shown in Figure 03, utilizes depth-wise separable convolutions, dividing a standard convolution into a depth-wise convolution and a point-wise convolution. Includes multiple layers of depth-wise separable convolutions with batch normalization and ReLU activation.

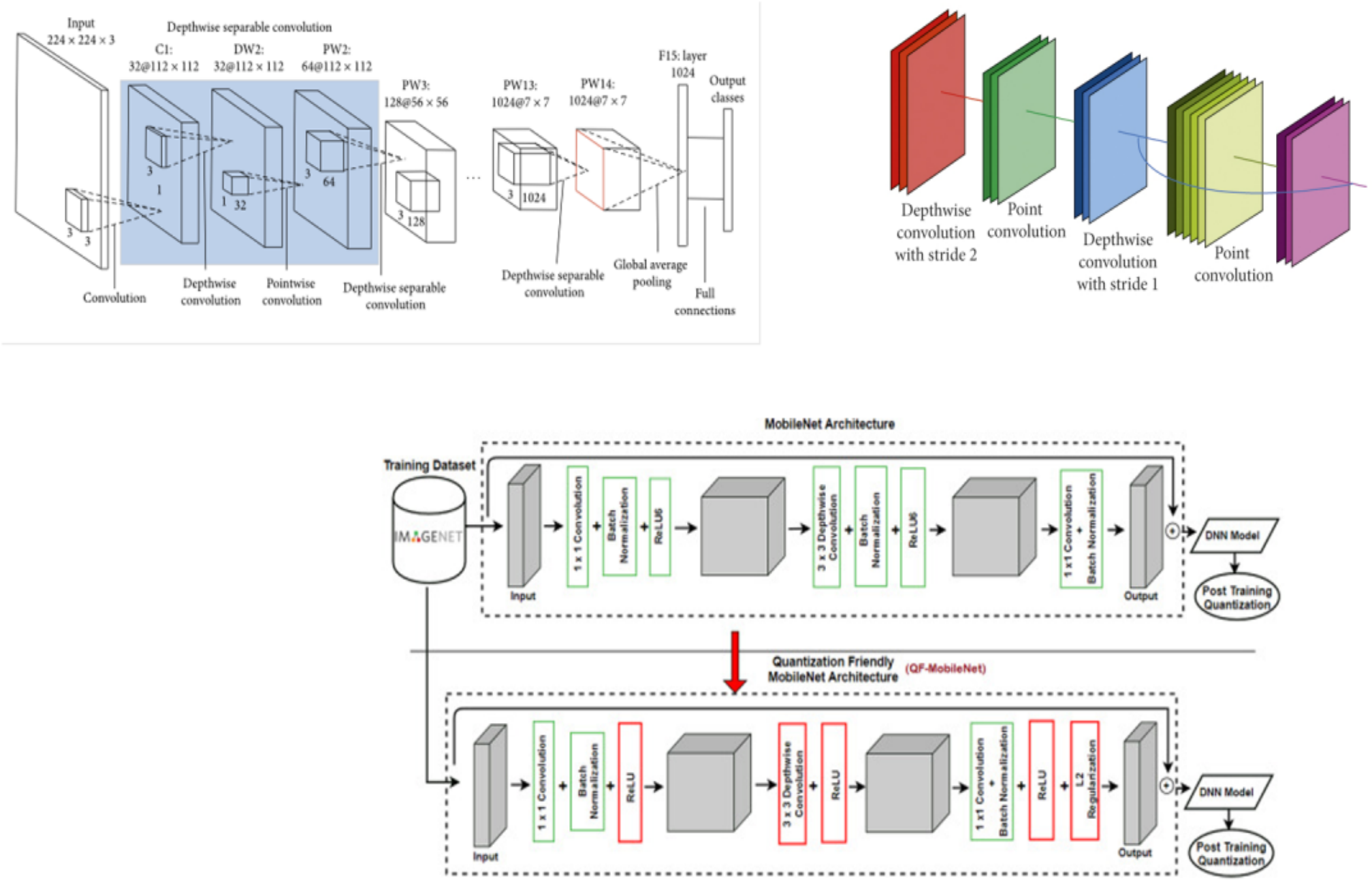

*Figure 2. 3: Architecture of MobileNet*

Performance:

- Accuracy: Competitive with larger models, though slightly lower due to its lightweight nature.
- Speed and Efficiency: Highly efficient, enabling real-time applications on mobile devices.
- Reasons for Efficiency: The use of depth-wise separable convolutions reduces both the computational cost and model size significantly without drastically reducing accuracy.

### 2.2.2 Inception v3

Inception v3[17] is a convolutional neural network architecture from the Inception family that achieves high accuracy on the ImageNet dataset with efficient computation.

The architecture includes features factorized convolutions and aggressive regularization. Uses multiple sizes of kernels in the same layer as shown in Figure 04 to capture information at various scales.

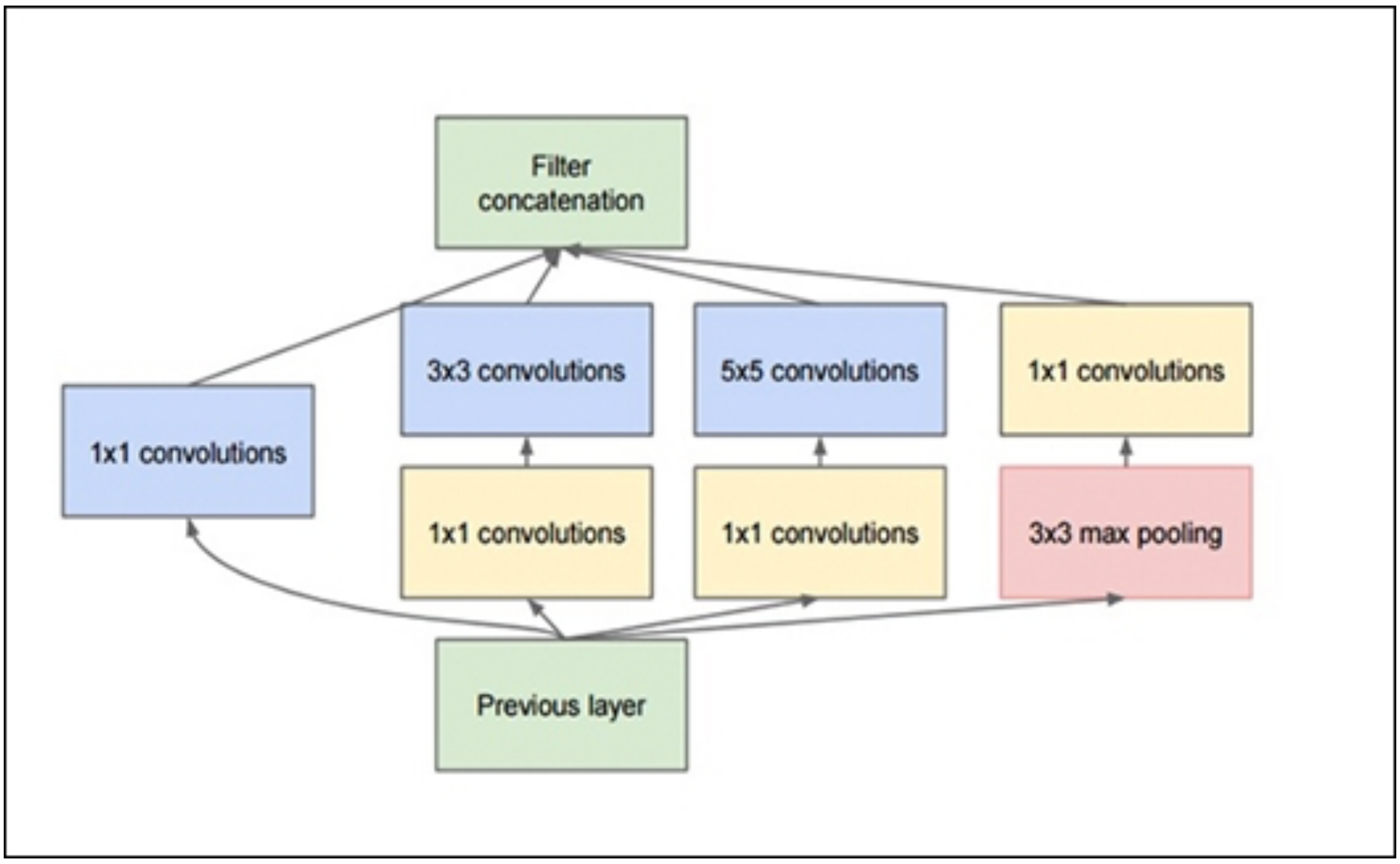

*Figure 2. 4: Architecture of Inception V3*

Performance:

- Accuracy: High accuracy on ImageNet and other datasets.
- Speed and Efficiency: More efficient than its predecessors due to factorization of convolutions and use of batch normalization.
- Reasons for Efficiency: The architecture cleverly reduces dimensionality and computational cost while maintaining high accuracy through factorized convolutions and extensive use of batch normalization.

### 2.2.3 VGG16 and VGG19

VGG16[18] and VGG19[19] are convolutional neural networks that are straightforward in architecture but deep, with 16 and 19 layers, respectively. Both consist of several convolutional layers with small 3×3 kernels, followed by max-pooling layers.

Fully connected layers followed by a softmax for output.
The difference between VGG16 which is shown in Figure 05- (a) and VGG19 which is shown in Figure 05- (b) is the depth; VGG19 has 3 more convolutional layers.

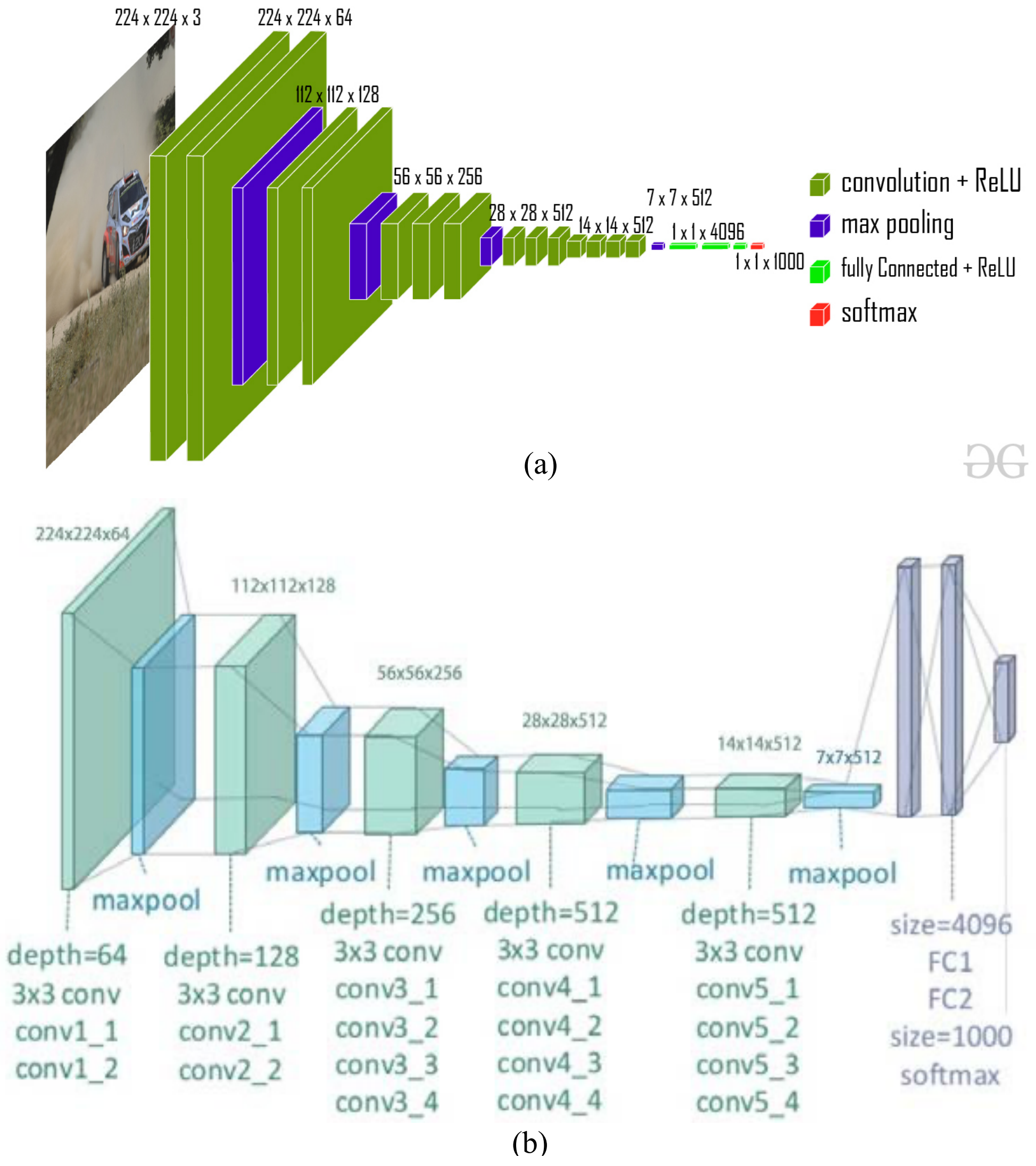

*Figure 2. 5: (a) - Architecture of VGG16, (b) - Architecture of VGG19*

Performance:

- Accuracy: Both achieve high accuracy on ImageNet, with VGG19 being slightly better due to its depth.
- Speed and Efficiency: These models are computationally intensive, making them slower compared to more modern architectures like MobileNet or Inception v3.

- Reasons for Efficiency and Speed: The simplicity and uniformity of the network's architecture allow for easy scaling but at the cost of increased computational load. The small kernel size used throughout the network allows for capturing fine details.

## 2.3 ROI Align (Region of Interest Align)

ROI Align (Region of Interest Alignment) [20] is a technique introduced as part of the Mask R-CNN framework for object detection and instance segmentation tasks in computer vision. It addresses the problem of aligning extracted features with input regions of interest (ROIs), improving upon previous methods like ROI Pooling used in Fast R-CNN. The goal of ROI Align is to maintain the spatial fidelity of features after down sampling, which is crucial for accurate object detection and segmentation.

ROI Align is a feature extraction technique that precisely aligns the extracted features with the input ROIs by avoiding quantization of ROI boundaries. Unlike ROI Pooling, which quantizes the ROI coordinates to the nearest integer before pooling (leading to misalignment between the ROI and the extracted features), ROI Align uses bilinear interpolation to compute the exact values of input features at four regularly sampled locations in each ROI bin, and then aggregates the samples via max or average pooling[21]. This shown in Figure 06- (a) and (b).

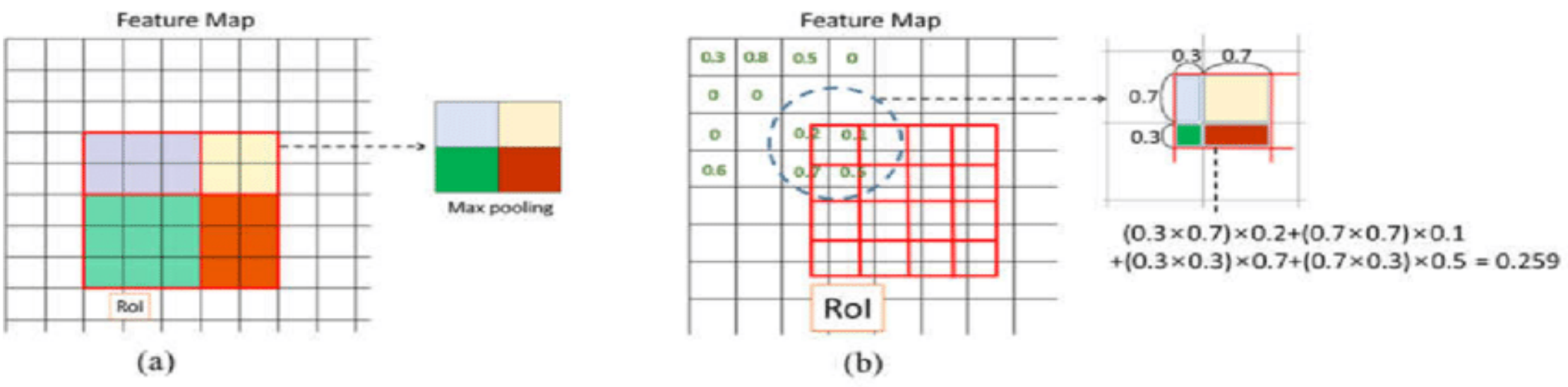

*Figure 2. 6: (a) ROI Pooling, (b) ROI Align*

### 2.3.1 Architecture of ROI Align

The architecture of ROI Align can be understood in the context of its operation within a Convolutional Neural Network (CNN) used for object detection or segmentation:

1. Feature Extraction: A CNN is used to extract a feature map from the input image.
2. ROI Proposal: A separate network proposes regions of interest (ROIs) that are likely to contain objects.
3. ROI Align Operation: For each ROI, the ROI Align layer:
    - Maps the ROI to the corresponding location on the feature map.
    - Divides the ROI into a fixed number of bins (e.g., a 7x7 grid).
    - For each bin, samples four points at regular intervals.
    - Uses bilinear interpolation to compute the feature value at each sample point.
    - Pools the sampled values (using max or average pooling) to form a single feature vector for each bin.
4. Classification and Regression: The pooled feature vectors are fed into fully connected layers to classify the object within the ROI and refine its bounding box.

## 2.3.2 Localizing an Image Using ROI Align

For localizing an image, ROI Align precisely maps the ROIs to the feature map, ensuring that the spatial information is accurately preserved. This process involves:

1. Sampling points within each ROI bin using bilinear interpolation to get precise feature values.
2. Pooling these values to reduce the dimensionality while maintaining spatial information.
3. Using the resulting feature vectors for accurate object localization and classification.

## 2.3.3 Advantages of ROI-align Over Other Localization Methods

- Improved Accuracy: By avoiding quantization errors and using bilinear interpolation, ROI Align maintains the spatial details of the features, leading to more accurate object localization and segmentation.
- Better Performance on Small Objects: The precise alignment helps in detecting small objects, where the spatial resolution of features is critical.
- Compatibility with Instance Segmentation: The improved feature alignment is particularly beneficial for instance segmentation tasks, where pixel-level accuracy is important.

## 2.3.4 Disadvantages of ROI Align

- Computational Complexity: The bilinear interpolation and sampling for each ROI bin can increase computational overhead, especially for large numbers of ROIs.

- Implementation Complexity: The process of bilinear sampling and pooling is more complex to implement compared to straightforward ROI Pooling.
- Sensitivity to ROI Proposals: The performance of ROI Align is still dependent on the quality of the ROI proposals. Poorly proposed ROIs can lead to suboptimal results.

### 2.3.5 Mathematical Foundations

The key mathematical operation in ROI Align is bilinear interpolation, used to compute the value of a feature at a non-integer pixel location. Given a point (x, y) and its four nearest pixel centres (x1, y1), (x2, y2), (x3, y3), (x4, y4) the value F(x, y) is computed as:

$$F(x, y) = \sum_{i=1}^{4} w_i \cdot F(x_i, y_i)$$

where wi are the weights calculated based on the distance of (x, y) to each of the four nearest pixel centres, ensuring that closer pixels contribute more to the interpolated value. This operation is performed for each of the sampled points in every ROI bin, followed by pooling to generate a fixed-size feature vector for each ROI, which can then be used for classification and bounding box regression.

## 2.4 GCN (Graph Convolutional Networks)

Graph Convolutional Networks (GCNs) [22] represent a significant advancement in the field of graph-based learning, where traditional neural network approaches struggle to capture the complexity of graph-structured data. Unlike images or text, which are inherently grid-like and sequential respectively, graphs embody a more intricate structure, consisting of nodes (vertices) and their interconnections (edges), making them suitable for representing complex relational data.

GCNs extend the concept of convolutional neural networks (CNNs) to graph data, allowing for the efficient aggregation and transformation of information from a node's neighbourhood. This process involves the application of convolutional operations that consider both the features of a node and the collective features of its neighbours, effectively capturing the local graph topology within each node's representation. This is explained in Figure 07.

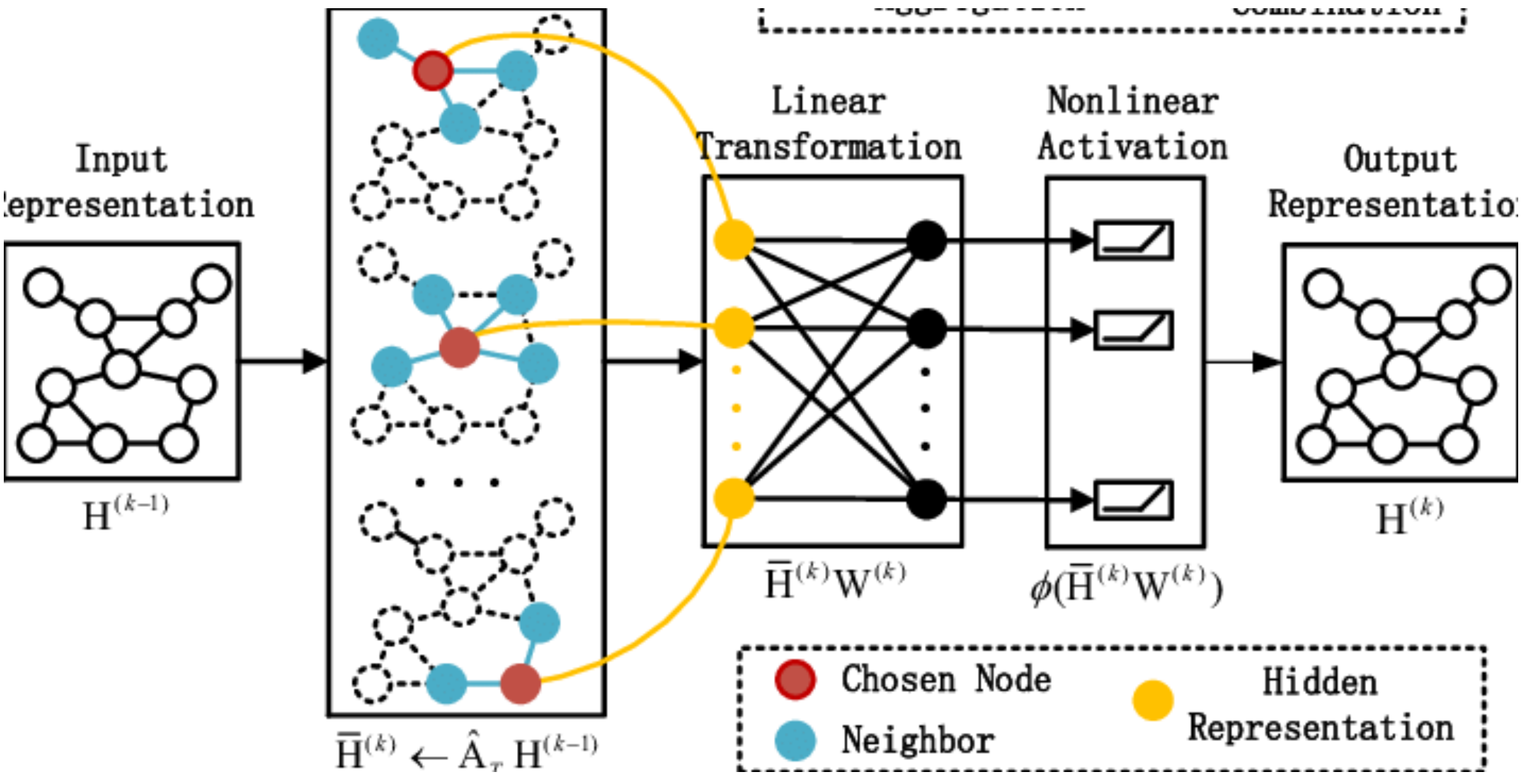

*Figure 2. 7: Architecture of GCN*

The power of GCNs lies in their ability to learn meaningful representations of nodes or entire graphs, depending on the task at hand, making them highly versatile. They have been successfully applied in various domains, including social network analysis, recommendation systems, natural language processing, and bioinformatics. By leveraging the rich relational information contained in graphs, GCNs facilitate a deeper understanding and modeling of complex systems, opening new avenues for research and application in machine learning and artificial intelligence.

## 2.5 ARG

The Actor Relation Graph (ARG)[23] is a novel approach designed to model the dynamic and complex relationships between actors in group activities within multi-person scenes. This methodology leverages deep models to efficiently learn discriminative relations between actors, utilizing both appearance and position information. By implementing Graph Convolutional Networks (GCN), ARG allows for the automatic learning of actor connections from video data in an end-to-end manner, significantly enhancing the efficiency and effectiveness of group activity recognition. The research demonstrates the superiority of ARG through extensive experiments on standard datasets, achieving state-of-the-art performance. Additionally, the ability of ARG to capture discriminative relation information crucial for accurately recognizing group activities is visually illustrated, as shown in Figure 08, showcasing its potential in advancing video understanding and behaviour analysis domains.

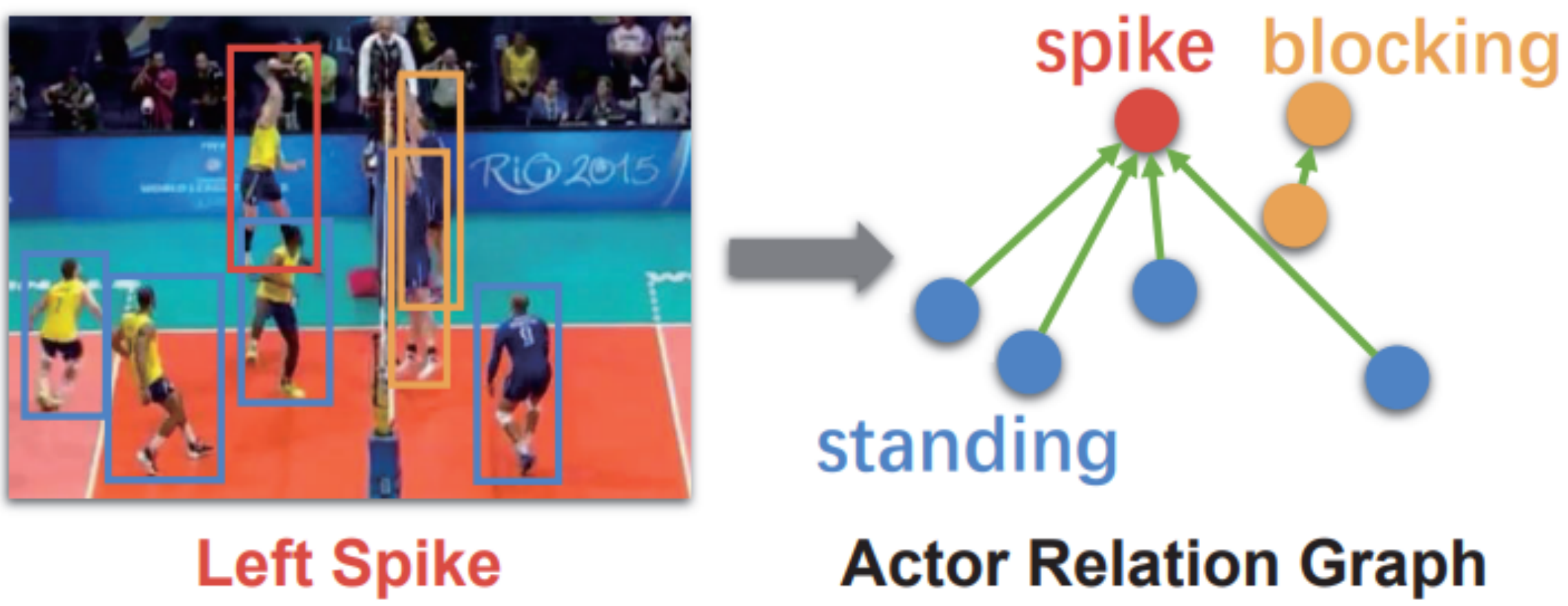

*Figure 2. 8: Actor Relation Graph for respective frames*

## 2.6 Similarity Searches for Appearances

In the context of computer vision and image processing, the concept of "appearance relation" refers to the methods and techniques used to analyse and determine the similarity or differences between different regions or features within an image or between different images. This concept is fundamental in various applications such as image matching, object recognition, tracking, and more. The appearance relation quantifies how closely areas of an image resemble each other or match with areas in another image, considering attributes like colour, texture, shape, and spatial arrangement.

To calculate the appearance relation, several methods can be employed, each with its mathematical foundation and application scenarios. Three widely used methods are Sum of Absolute Differences (SAD) [24], Normalized Cross-Correlation (NCC)[25], and Dot Product[26]. Below, we delve into each method, exploring its mathematical underpinnings and providing examples for clarity.

### 2.6.1. Sum of Absolute Differences (SAD)

The Sum of Absolute Differences is a simple yet effective method for comparing similarity between two image regions. It calculates the absolute difference between corresponding pixels in the two regions and sums these differences. Mathematically, it can be expressed as:

$$\mathrm{SAD}(I_1, I_2) = \sum_{x,y} |I_1(x, y) - I_2(x, y)|$$

where $I_1$ and $I_2$ represent two image regions, and $x$, $y$ are the coordinates of pixels within these regions.

### 2.6.2. Normalised Cross-Correlation (NCC)

NCC is a measure of similarity between two image regions that accounts for differences in lighting or exposure by normalising the data. It is defined as:

$$\mathrm{NCC}(I_1, I_2) = \frac{\sum_{x,y}(I_1(x,y)-\bar{I}_1)(I_2(x,y)-\bar{I}_2)}{\sqrt{\sum_{x,y}(I_1(x,y)-\bar{I}_1)^2 \sum_{x,y}(I_2(x,y)-\bar{I}_2)^2}}$$

where $\bar{I}_1$ and $\bar{I}_2$ are the mean intensity values of the image regions $I_1$ and $I_2$, respectively.

### 2.6.3. Dot Product

The dot product between two vectors is a measure of their similarity in direction and magnitude. In the context of appearance relation, it can be used to compare the feature vectors extracted from image regions. Mathematically, the dot product is defined as:

$$\text{Dot Product}(V_1, V_2) = \sum_{i=1}^{n} V_{1i} \cdot V_{2i}$$

where $V_1$ and $V_2$ are feature vectors extracted from two image regions, and $n$ is the dimension of the vectors.

Understanding and calculating the appearance relation between image regions is crucial for various applications in computer vision. Methods like SAD, NCC, and the dot product offer different perspectives and mathematical frameworks to quantify these relations. Each method has its strengths and application scenarios, from simple difference calculations with SAD to complex similarity measures with NCC and vector-based comparisons with the dot product. Choosing the appropriate method depends on the specific requirements of the task, including sensitivity to lighting conditions, computational efficiency, and the nature of the features being compared.

## 2.7 Image Segmentation

Image segmentation is a crucial task in computer vision that involves dividing an image into meaningful parts or segments, making it easier to analyze or process. Each segment typically corresponds to a different object or region within the image, allowing for more detailed image understanding. The primary goal of image segmentation is to simplify the representation of an image and make it more meaningful for specific tasks, such as object detection, scene understanding, or image editing.

Image segmentation can be classified into different types:

1. Semantic Segmentation: Each pixel in the image is classified into a particular class. For example, in a street scene, pixels could be classified as "road," "car," "building," etc.

2. Instance Segmentation: Extends semantic segmentation by distinguishing between different instances of the same class. For example, in a crowd scene, multiple people would be segmented as distinct instances.

3. Panoptic Segmentation: Combines semantic and instance segmentation, assigning a class and instance label to each pixel.

## 2.7.1 Models Used for Image Segmentation

### I. U-Net

U-Net is a neural network designed for image segmentation, particularly in biomedical applications. Its architecture consists of an encoder (downsampling path) that captures context and a symmetric decoder (upsampling path) that recovers spatial details. The "U" shape comes from the skip connections that link corresponding layers in the encoder and decoder, allowing the model to combine high-level features with spatial information. This design helps U-Net generate accurate segmentation maps, even with limited training data, making it ideal for tasks like identifying cells in medical images.

### II. Mask R-CNN

Mask R-CNN extends the Faster R-CNN model by adding a branch for predicting segmentation masks on each detected object, in addition to bounding boxes and class labels. It works by first detecting objects using a region proposal network (RPN) and then applying a mask head to produce a pixel-level binary mask for each object instance. This model excels in instance segmentation, where the goal is to identify and separate different objects within the same category, like distinguishing between individual people in a crowd.

### III. DeepLab

DeepLab is a segmentation model that introduces atrous (dilated) convolutions to extract dense feature maps without reducing spatial resolution, enabling the model to capture fine details. Additionally, it incorporates a Conditional Random Field (CRF) to refine the segmentation output, particularly around object boundaries. DeepLab can handle multi-scale features and complex scenes, making it effective for semantic segmentation tasks such as labeling objects in street scenes. Variants like DeepLabV3+ have further improved accuracy by integrating encoder-decoder architectures and more sophisticated feature extraction methods.

### IV. SegNet

SegNet is an encoder-decoder architecture designed for semantic segmentation tasks. The encoder part is similar to the convolutional layers of the VGG16 network, which reduces the spatial resolution of the input image while capturing feature information. The decoder part upsamples the low-resolution feature maps back to the original image size, predicting pixel-wise class labels. What sets SegNet apart is its efficient use of memory, as it only stores the indices of the max-pooling layers during encoding, making it suitable for real-time applications like autonomous driving.

### 2.7.2 Comparison of U-Net, Mask R-CNN, DeepLab, and SegNet

| Model | Purpose | Architecture | Strengths | Weaknesses |
|---|---|---|---|---|
| U-Net | Semantic segmentation | U-shaped encoder-decoder | High accuracy for biomedical images, skip connections | Not suitable for object detection, limited scalability |
| Mask R-CNN | Object detection & instance segmentation | Extended Faster R-CNN with segmentation branch | Simultaneous object detection & segmentation, versatile | Computationally expensive, complex training |
| DeepLab | Semantic segmentation | Atrous convolutions & ASPP | Captures multi-scale context, accurate boundary detection | High computational cost, complex architecture |
| SegNet | Semantic segmentation | Encoder-decoder with index pooling | Efficient, good for real-time applications | Lower accuracy compared to U-Net, limited to semantic segmentation |

*Table 2. 1: Comparison of U-Net, Mask R-CNN, DeepLab, and SegNet*

Mask R-CNN is superior for tasks requiring both bounding box coordinates and precise masking coordinates because it seamlessly integrates object detection and instance segmentation within a single framework. Unlike models that focus solely on segmentation, Mask R-CNN not only identifies and localizes objects by generating accurate bounding boxes but also produces pixel-level masks that capture the exact shape of each object. This dual capability ensures that Mask R-CNN provides detailed and accurate spatial information, making it particularly effective for applications where both object localization and fine-grained segmentation are critical.

### 2.7.2 Mask-RCNN for image segmentation

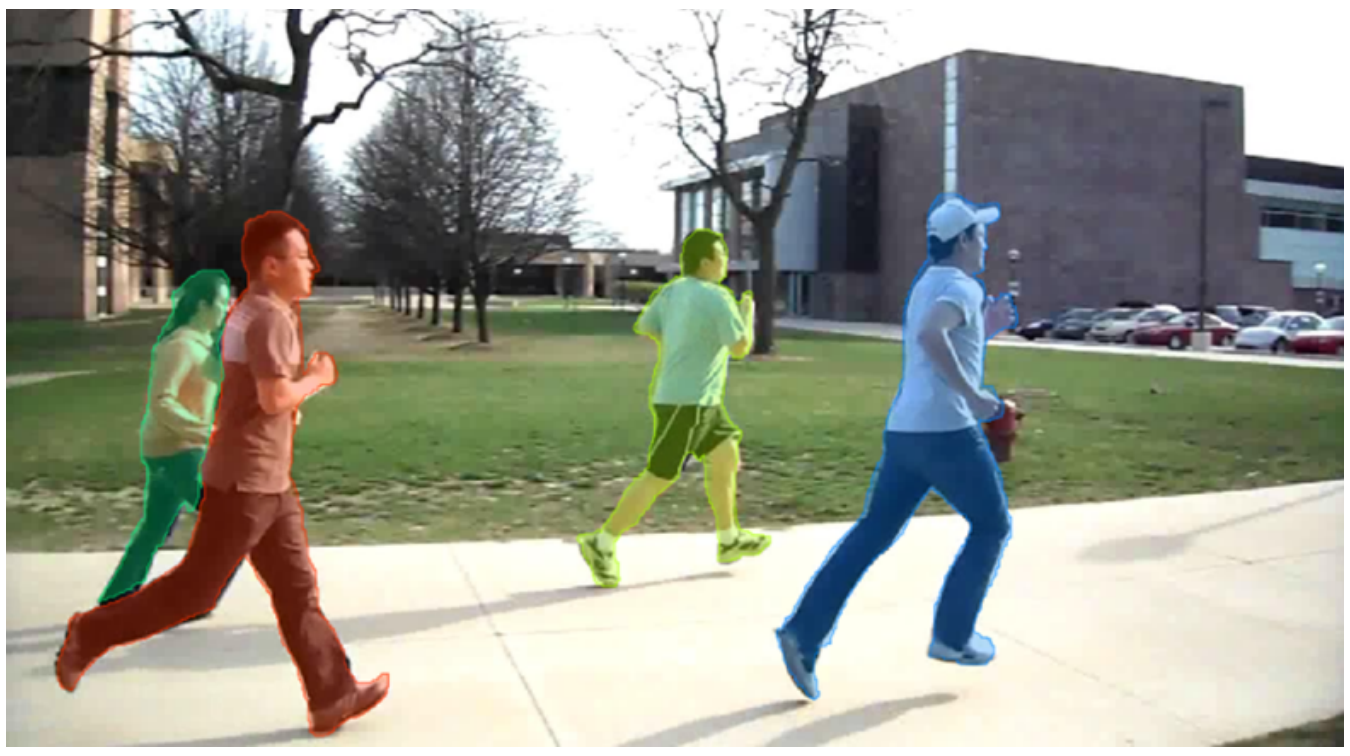

*Figure 2. 9: Image segmentation with Mask-RCNN*

Mask R-CNN is a powerful deep learning framework that excels in tasks requiring precise object localization and segmentation. One of its prominent applications is in people detection and segmentation in images or videos, as shown in Figure 09, a task crucial for a variety of domains including surveillance, autonomous driving, and human-computer interaction.

- **Architecture of Mask R-CNN**

Mask R-CNN is an advanced deep learning framework designed for object detection and instance segmentation tasks. It builds on the Faster R-CNN architecture by adding a branch for predicting segmentation masks in addition to the existing branches for bounding box detection and classification. This architectural enhancement enables Mask R-CNN to perform pixel-level segmentation, which is crucial for tasks requiring detailed object delineation. The architecture is show in Figure 10.

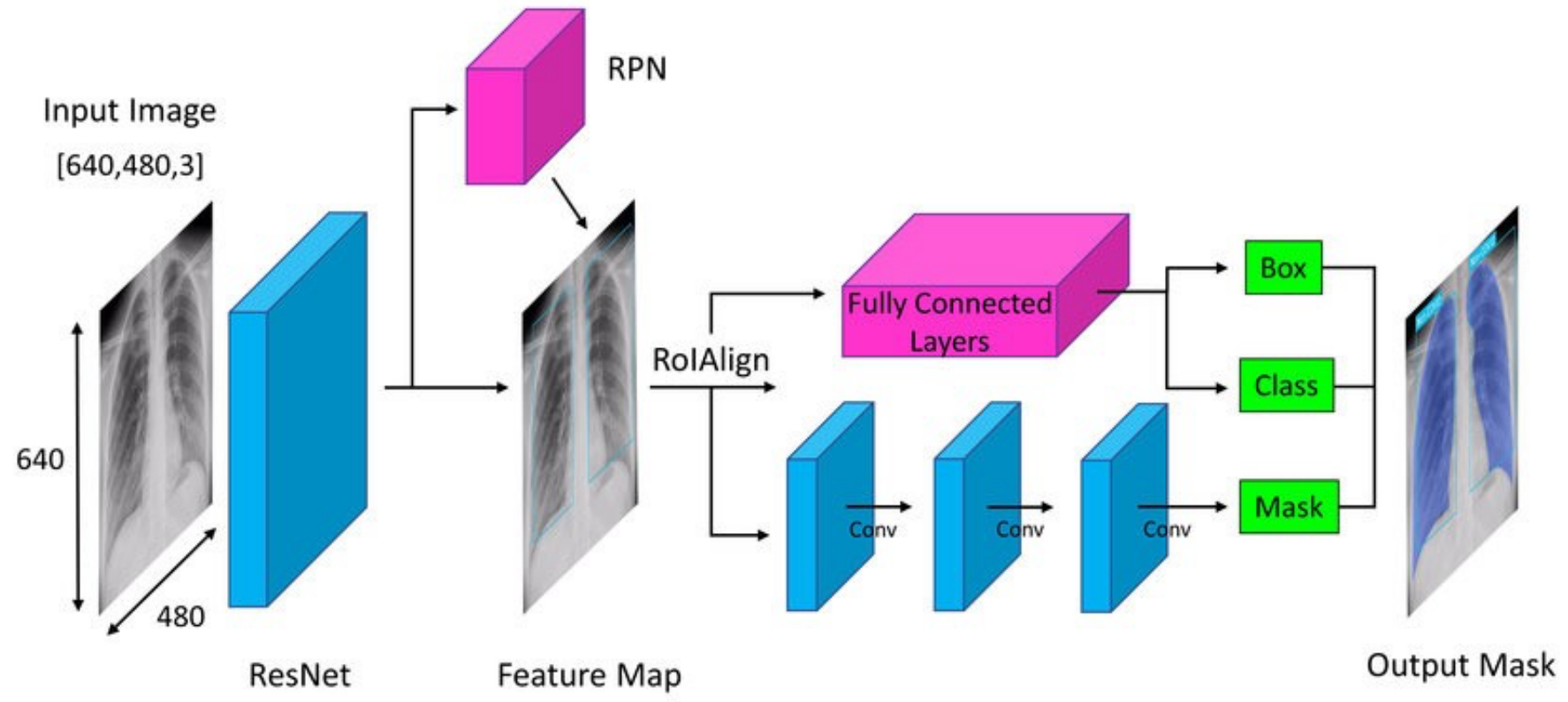

*Figure 2. 10: Mask RCNN architecture*

- **Backbone Network for Feature Extraction**

At the core of Mask R-CNN is the backbone network, typically a Convolutional Neural Network (CNN) pre-trained on large datasets like ImageNet. Commonly used backbones include ResNet, ResNeXt, and VGG. The backbone is responsible for extracting feature maps from the input image. These feature maps capture essential details such as edges, textures, and other spatial features that are crucial for detecting and segmenting objects.

ResNet (Residual Networks) is often chosen as the backbone due to its ability to train deep networks effectively. It introduces residual connections that help in mitigating the vanishing gradient problem, making it easier to train very deep networks.
Feature Pyramid Network (FPN) is sometimes used in conjunction with ResNet to create a feature pyramid, which allows the model to detect objects at multiple scales. The FPN generates feature maps at different levels of the pyramid, helping the model to be more robust to varying object sizes.

- **Region Proposal Network (RPN)**

The Region Proposal Network (RPN) is a crucial component that identifies potential regions of interest (RoIs) where objects might be located. The RPN is a lightweight network that slides over the feature maps generated by the backbone and predicts candidate bounding boxes.

- **Anchor Boxes**

The RPN generates anchor boxes, which are predefined bounding boxes of various sizes and aspect ratios. Each anchor box is evaluated to determine whether it likely contains an object.
Classification and Regression: The RPN outputs two types of information for each anchor box: a classification score that indicates whether the anchor contains an object, and a regression offset that refines the anchor box's coordinates to better fit the object.

- **RoI Align**

One of the key improvements Mask R-CNN introduces over Faster R-CNN is RoI Align. RoI Align addresses the spatial misalignment issues caused by RoI Pooling, which uses quantization operations that can lead to inaccuracies in segmentation tasks.

Bilinear Interpolation: Instead of quantizing the coordinates to the nearest grid point, RoI Align uses bilinear interpolation to compute the exact values of the feature maps at each RoI. This approach preserves spatial information more accurately, leading to better mask predictions.

- **Bounding Box Regression and Classification**

After RoI Align, the aligned features are processed by fully connected layers that perform two tasks:

  - **Bounding Box Regression**: The network predicts the precise coordinates of the bounding box for each RoI. These coordinates refine the initial proposals made by the RPN, ensuring that the bounding box tightly encloses the object.
  - **Classification**: The network also predicts the class label for each RoI, determining which object class (e.g., person, car, dog) is present within the bounding box.

- **Mask Branch for Instance Segmentation**

The mask branch is the distinctive feature of Mask R-CNN. This branch is a fully convolutional network (FCN) that operates in parallel with the classification and bounding box regression branches. It predicts a binary mask for each RoI, indicating which pixels within the bounding box belong to the object.

Mask Prediction: The mask branch outputs a mask of fixed size (e.g., 28x28 pixels) for each object class. Each mask is a binary array where each pixel value indicates whether the corresponding pixel in the original image is part of the object.
Per-Class Masking: A separate mask is predicted for each object class. During inference, only the mask corresponding to the predicted class is used, ensuring that the mask is aligned with the detected object.

- **Loss Function and Training**

Mask R-CNN uses a multi-task loss function that combines the losses from the classification, bounding box regression, and mask prediction tasks. This loss function is designed to optimize all three tasks simultaneously, leading to a model that is highly accurate in detecting and segmenting objects.

Classification Loss: This is typically a cross-entropy loss that measures the accuracy of the predicted class labels.
Bounding Box Loss: This is usually a smooth L1 loss that measures how well the predicted bounding box coordinates match the ground truth.
Mask Loss: This is a pixel-wise binary cross-entropy loss applied only to the pixels within the bounding box, optimizing the predicted mask's accuracy.

# CHAPTER 3: METHODOLOGY

## 3.1 Introduction

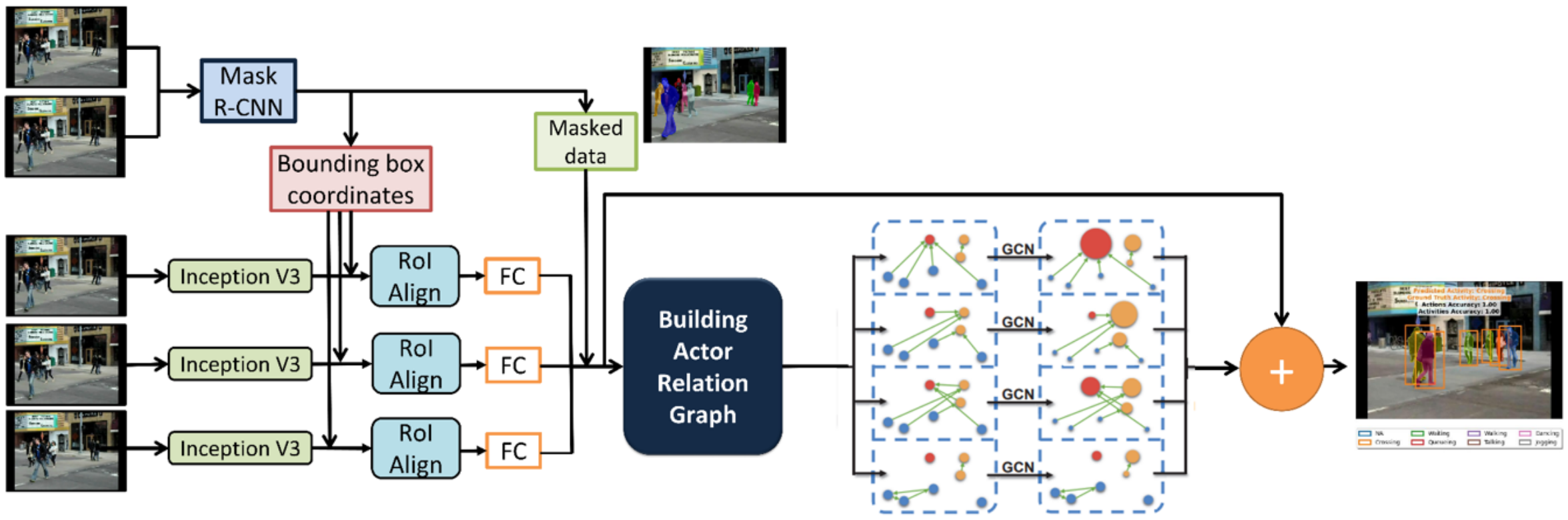

*Figure 3. 1: Overview of our Network*

Our project builds upon a cutting-edge approach in the field of computer vision and machine learning for recognizing group activities in multi-person scenes. The foundational method involves, as elaborated in Figure 11, constructing Actor Relation Graphs (ARG) to represent the dynamics within multi-person scenes and applying Graph Convolutional Networks (GCN) to perform relational reasoning for activity recognition. This methodology, derived from recent advances in relational reasoning and graph neural networks, serves as the backbone of our project.

The core framework of our approach begins with the extraction of feature vectors from video sequences, utilizing bounding boxes and mask data to identify and localize actors within a scene. Originally, this process employed the Inception-v3 architecture for feature extraction, combined with Roi Align for precise localization of actor features within the frame. Building upon this, ARGs are constructed to model the pairwise relationships between actors, integrating both appearance and positional information to capture the complex dynamics of group activities.

Our project extends this methodology by incorporating Mask R-CNN to enhance the accuracy and precision of actor localization and feature extraction. Mask R-CNN is utilized for two key objectives: first, to extract mask data for actors in the images, providing detailed segmentation of individuals; second, to determine the exact boundary box coordinates for these actors. The boundary box coordinates extracted from Mask R-CNN are used to filter the feature map of actors in the scene, and the mask data is convolved with the boundary box feature map to create a masked feature map for each actor across a sequence of images. This enhanced feature map improves the representation of each actor, enabling more accurate analysis of their interactions.

Recognizing the impact of scene complexity and environmental conditions on activity recognition performance, we implemented a comparative analysis across different scenarios, such as crowded and less crowded scenes.

- **Feature Extraction Backbones**: Initially, the project adopted the Inception-v3 model for feature extraction. However, to evaluate the effectiveness of different architectures, we iteratively experimented with alternative backbones, including MobileNet and VGG16. These changes aimed to assess the influence of different feature extraction models on the system's ability to accurately capture actor features, especially when combined with the masked feature maps generated by Mask R-CNN.

- **Mask R-CNN Implementation**: To improve the precision of actor localization and feature extraction, Mask R-CNN was integrated into our methodology. This model not only provided the bounding box coordinates but also generated mask data for each actor. By convolving the mask data with the feature maps, we created masked feature maps that better represented individual actors within the scene, leading to more accurate group activity detection.

- **Appearance Similarity Search Algorithms**: To refine the construction of ARGs, we initially utilized the Normalized Cross Correlation (NCC) for evaluating appearance similarities between actors. We also explored alternative algorithms, including Sum of Absolute Differences (SAD) and dot-product similarity, to enhance relational modeling between actors. This exploration was particularly important in complex scenes where actor interactions are densely interconnected or obscured by environmental factors.

- **Scenario-Specific Adaptations**: Each scenario—crowded scenes and less crowded scenes—presents unique challenges for group activity recognition. Our methodology includes tailored adaptations to address these challenges, such as adjusting the feature extraction approach and similarity search mechanism to better accommodate the specific characteristics of each scenario. The use of Mask R-CNN in crowded scenes, for example, ensures more accurate segmentation and localization of actors, even in densely populated environments.

## 3.2 Selected Architectures

For our research project, focused on achieving high accuracy in computer vision-based group activity detection, we meticulously evaluated various architectures before finalizing our approach. The chosen methodologies underscore our commitment to precision and detail in feature extraction, localization, and activity recognition.

We opted for Convolutional Neural Networks (CNNs)—specifically Inception v3, MobileNet, and VGG16—coupled with Mask R-CNN, over YOLO (You Only Look Once). This decision was driven by the superior feature extraction and precise localization capabilities of these CNNs and Mask R-CNN. Unlike YOLO, which is designed for speed and efficiency in object detection, Inception v3, MobileNet, and VGG16 provide a richer and more detailed analysis of the visual content. These networks, with their varying depths and architectures, are adept at capturing a wide array of features at different scales, essential for accurately identifying and analyzing the nuanced aspects of group activities. This level of detail is crucial for our project's focus on accuracy.

The integration of Mask R-CNN into our framework presents significant advantages over traditional bounding box approaches. Mask R-CNN not only provides accurate boundary box coordinates but also generates detailed mask data for each actor, allowing for precise segmentation within the video frames. This mask data is then convolved with the feature maps extracted by the CNNs, resulting in masked feature maps that offer a more refined representation of each actor. This approach enhances the accuracy of actor localization and ensures that only the relevant features are analyzed, which is crucial for the subsequent group activity detection.

The decision to use Mask R-CNN over simpler methods like RoIAlign was based on its ability to address the limitations of bounding box-based localization. While RoIAlign improves upon traditional methods by reducing quantization error, Mask R-CNN goes a step further by offering both precise localization and segmentation, which are critical for accurately capturing the complex dynamics within group activities. This dual capability makes Mask R-CNN particularly well-suited for our project's needs.

For group activity detection, we selected Graph Convolutional Networks (GCN) over alternatives such as 3D-CNN, two-stream CNNs, and CNN combined with LSTM. GCNs were chosen due to their superior ability to model the complex interactions and relationships between actors in a scene. Unlike the other architectures considered, which primarily focus on extracting spatial or temporal features independently, GCNs allow for effective relational reasoning through the structure of Actor Relation Graphs (ARGs). This graph-based approach is ideal for analyzing group activities, where understanding the interactions and dynamics between participants is critical for accurate recognition.

By combining advanced feature extraction using CNNs with the precision of Mask R-CNN for localization and segmentation, and leveraging GCNs for relational reasoning, our architecture is designed to achieve high accuracy in group activity detection. This carefully selected combination of methodologies ensures that our system can effectively capture, analyze, and interpret the complex interactions within multi-person scenes.

## 3.3 Dataset

In our research, we used the public group activity recognition datasets called collective activity dataset[] and augmented dataset [] to train and test our model. This dataset has 74 video scene that includes multiple persons in each scene. The manually defined bounding boxes on each person and the ground truth of their actions and the group activity are also labelled in each frame which is shown in Figure 12.

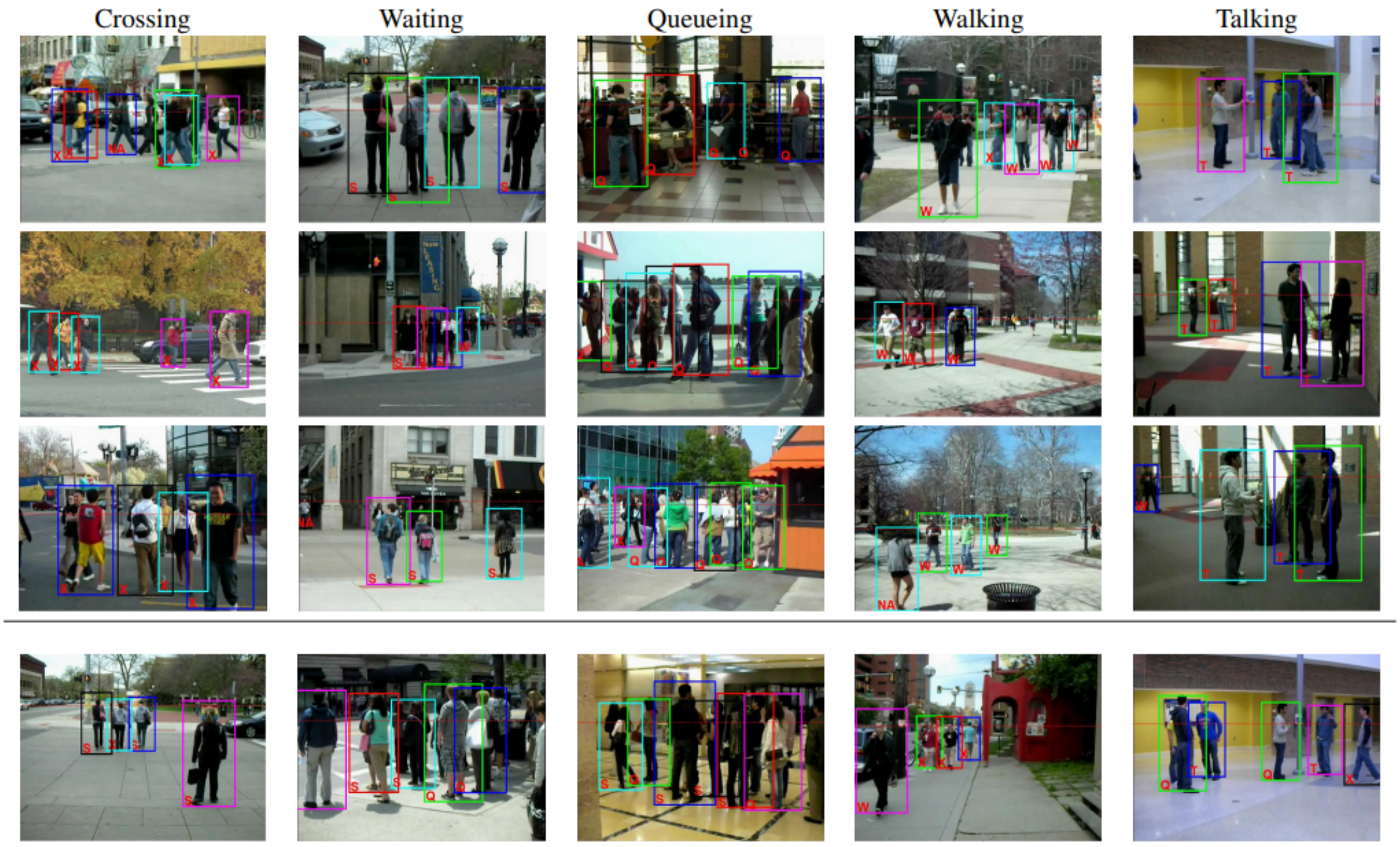

*Figure 3. 2: Collective Activity Dataset*

## 3.4 Boundary box coordinates extraction and mask generation

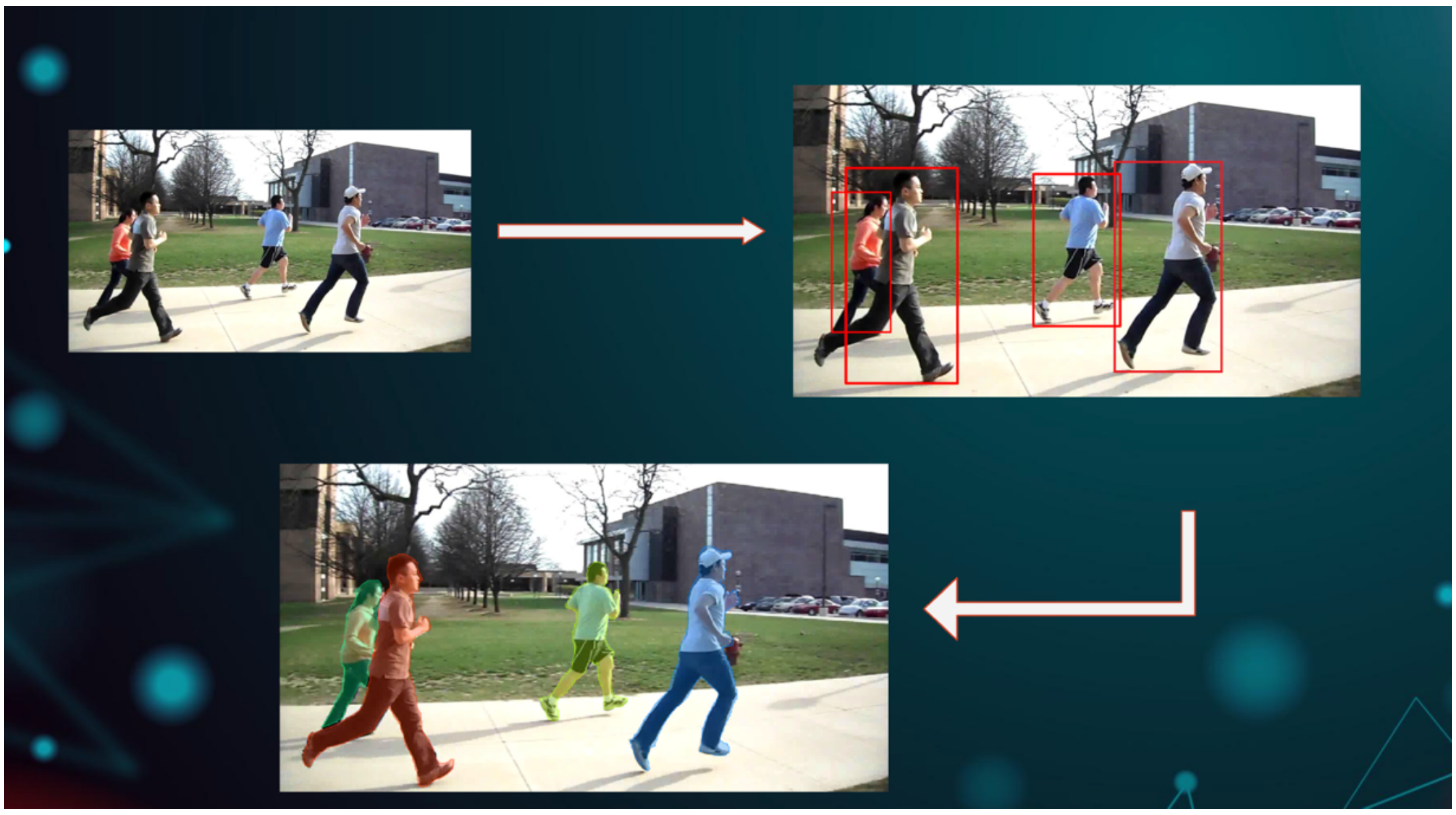

*Figure 3. 3: Bounding box and Mask generation*

In our methodology, Figure 13 describes the process of boundary box coordinates extraction and mask generation is pivotal for accurately localizing and representing the features of individuals within video frames. This section outlines the step-by-step process, integrating key architectures such as ResNet, Region Proposal Networks (RPN), Mask R-CNN, and CNNs like Inception v3, MobileNet, and VGG16 to achieve precise feature extraction and segmentation.

### 3.4.1 Boundary Box Coordinates Extraction

- **Feature Extraction Using ResNet**: The first step in our pipeline involves extracting initial features from the input images using a ResNet (Residual Network) architecture. ResNet is chosen for its ability to maintain high accuracy while mitigating the vanishing gradient problem, which allows for deeper network construction. Specifically, we use ResNet as the backbone network within the Mask R-CNN framework.

- **Region Proposal Network (RPN)**: The extracted features from ResNet are then fed into a Region Proposal Network (RPN). The RPN is responsible for generating candidate regions (region proposals) that are likely to contain objects (in our case, actors). The RPN commonly used in Mask R-CNN is based on a fully convolutional network (FCN) that predicts object bounds and objectness scores at each position in the feature map. It works by sliding a small network over the convolutional feature map output by ResNet, producing a set of anchor boxes with varying scales and aspect ratios. For each anchor box, the RPN generates two outputs: an objectness score (indicating the likelihood of an object being present) and a bounding box regression (refining the anchor to better fit the object).

- **Bounding Box Refinement**: The RPN outputs a set of bounding boxes, which are further refined using a process called non-maximum suppression (NMS). NMS eliminates redundant and overlapping boxes by selecting the box with the highest objectness score and discarding others that overlap significantly with it. The final output of the RPN is a set of precise bounding box coordinates for the actors within the video frames.

### 3.4.2 Integration with Feature Extraction

- **Feature Extraction Using Inception v3**: Concurrently, the input images undergo another round of feature extraction using Inception v3, a CNN architecture known for its ability to capture high-level features across multiple scales. This step is crucial for generating a rich feature map that will later be used for accurate activity recognition.

- **ROIAlign for Feature Map Refinement**: The bounding box coordinates extracted by the RPN are then integrated with the feature map generated by Inception v3. This integration is done using a technique called ROIAlign. ROIAlign accurately maps the bounding box coordinates onto the feature map, ensuring that the features within the bounding box are precisely aligned with the original image pixels. This eliminates the quantization errors that can occur with other pooling methods like RoIPool. The result is a refined feature map that accurately represents the human features within the localized bounding boxes.

### 3.4.3 Mask Generation and Final Feature Map Creation

- **Mask Generation Using Mask R-CNN**: Mask R-CNN extends the object detection capabilities of the Faster R-CNN by adding a branch for predicting segmentation masks. In our framework, after the bounding box coordinates are refined using ROIAlign, Mask R-CNN generates a binary mask for each actor. This mask represents the exact shape of the actor within the bounding box, providing pixel-level segmentation.
- **Convolution of Mask and Feature Map**: The binary masks generated by Mask R-CNN are then convolved with the refined feature maps obtained from ROIAlign. This convolution process enhances the feature map by applying the mask data, ensuring that the final feature map focuses exclusively on the features corresponding to the human actors, filtering out irrelevant background information.
- **Final Feature Map Construction**: The final output is a highly accurate feature map that combines the rich feature extraction capabilities of CNNs (Inception v3, MobileNet, VGG16) with the precise localization and segmentation provided by Mask R-CNN. This feature map serves as the input for further processing, such as constructing Actor Relation Graphs (ARGs) and applying Graph Convolutional Networks (GCNs) for group activity detection.

## 3.5 Feature Extraction and Localization

Our project employs a sophisticated feature extraction and localization methodology, pivotal for analyzing group activities in video sequences. This methodology integrates the strengths of Convolutional Neural Networks (CNNs), specifically Inception v3, MobileNet, and VGG16, alongside RoIAlign for precise feature localization. Here, we delve into the mathematical operations and architectural nuances of these models to elucidate their contributions to our research.

### 3.5.1 Convolutional Neural Networks (CNNs)

At the core of CNNs lies the convolutional operation, a mathematical process designed to extract patterns from images.

This operation is defined as,

$$F_{out}(x, y) = \sum_{i=-a}^{a} \sum_{j=-b}^{b} K(i, j) \cdot F_{in}(x - i, y - j)$$

Where,

| | |
|---|---|
| $F_{out}$ | - Output feature map, |
| $F_{in}$ | - Input image or feature map, |
| K | - Kernel or filter, |
| (x,y) | - Spatial coordinates. |

This process enables the network to capture essential visual features like edges, textures, and patterns at various spatial hierarchies.

**Inception v3**

Inception v3 introduces a mix of filters of varying sizes within the same layer, allowing it to capture information across different scales. It employs a module-based architecture, where each module comprises parallel convolutional layers with filters of sizes 1x1, 3x3, and 5x5. This diversity in filter sizes enables the network to adaptively focus on both local and global features within the input image. The concatenation of these parallel layers' outputs allows for a rich, multi-scale feature representation.

**MobileNet**

MobileNet utilizes depthwise separable convolutions, which factorize a standard convolution into a depthwise convolution and a 1x1 pointwise convolution. This approach reduces computational complexity significantly. Mathematically, this can be expressed as:

$$F_{out} = (F_{in} * K_{depthwise}) * K_{pointwise}$$

where,

$K_{depthwise}$ - Performs lightweight filtering by applying a single filter per input channel

$K_{pointwise}$ - A 1x1 convolution, combines the outputs of the depthwise layer.

This efficient architecture ensures rapid processing without sacrificing feature extraction quality.

**VGG16**

VGG16's architecture is characterized by its simplicity, employing small 3x3 convolutional filters in a deep sequential arrangement. This uniformity allows for an incremental build-up of features and contributes to the network's robustness in capturing complex patterns. The mathematical essence of VGG16's convolutional layers can be summarized by the repeated application of the convolutional operation, enabling deep feature hierarchies to be constructed.

## 3.5.2 Localization with RoIAlign

RoIAlign corrects the misalignments caused by the coarse spatial quantization of RoIPool, offering precise feature extraction for object detection tasks. RoIAlign eliminates the harsh floor operation found in RoIPool by using bilinear interpolation to compute the exact values of input features at four sampling points in each RoI, followed by aggregating these values (usually by max or average pooling). This method can be mathematically represented as:

$$F_{roi} = Pool(Interpolate(F_{in}, S))$$

\

Where,

S - The set of sampled points within each RoI.

By this means, RoIAlign ensures that the extracted features are accurately aligned with the input, enhancing the model's performance in recognizing and analyzing group activities.

In conclusion, our methodology for feature extraction and localization is meticulously designed to harness the power of advanced CNN architectures and precise alignment techniques for analyzing video sequences. By uniformly sampling a set of K frames from the video, we ensure a comprehensive representation of the scene across time. Our use of Inception-v3 for multi-scale feature map extraction from each frame, supplemented by experiments with other backbone models like MobileNet and VGG16, demonstrates our commitment to identifying the most effective approach for capturing the nuanced details of actor interactions within a scene. The application of RoIAlign for extracting features from actor bounding boxes ensures that these details are accurately localized, preserving spatial integrity. This process culminates in the construction of a $N \times d$ matrix X, which concisely represents the feature vectors of actors across the sampled frames. This matrix forms the foundation of our subsequent analysis, enabling the sophisticated interpretation of group activities through the relational dynamics of the actors involved. Through this approach, we aim to achieve unparalleled accuracy in computer vision-based group activity detection, setting a new benchmark for the field.

## 3.6 Building Actor Relation Graphs

Our project enhances group activity recognition through a detailed methodology for constructing Actor Relation Graphs (ARGs), integrating both mathematical formulations and algorithmic strategies. This dual approach allows for a comprehensive analysis of the pair-wise relationships between actors in a scene, leveraging appearance and positional information for a nuanced understanding of group dynamics.

**Graph Structure and Computational Framework**

We define the ARG where each node represents an actor within the scene, formalized as,

$$A = \{(x_a^i, x_s^i) | i = 1, \ldots, N\}$$

Here, N signifies the number of actors,
$x_a^i$ encodes the appearance features in a d-dimensional space, and $x_s^i$ represents the spatial coordinates of each actor's bounding box. The relational matrix $G \in R^{N \times N}$ quantifies the importance of each actor's features to every other actor, encapsulating the complex web of interactions within the scene.

The crux of our ARG construction lies in accurately calculating the relation values,
$G_{ij}$, which integrate both the appearance and positional relations through a composite function,

$$G_{ij} = \frac{fs(x_s^i, x_s^j) \exp(fa(x_a^i, x_a^j))}{\sum_{j=1}^{N} fs(x_s^i, x_s^j) \exp(fa(x_a^i, x_a^j))}$$

This formula ensures that relations are normalized across each node, enabling a balanced representation of actor interactions.

### 3.6.1. Appearance Relation Techniques

To capture the essence of actor appearance, we employ three distinct computational methods, each bringing a unique perspective on similarity and relation:

**1. Embedded Dot-Product**: Inspired by attention mechanisms, this approach computes similarity in an embedding space, enhancing the model's ability to discern nuanced appearance traits. The transformation functions θ and ϕ project appearance features into a subspace, facilitating a more refined relation analysis.

$$fa(x_a^i, x_a^j) = \frac{\theta(x_a^i)^T \phi(x_a^j)}{\sqrt{d_k}}$$

**2. Normalized Cross-Correlation (NCC)**: NCC provides a robust measure of similarity, accounting for variations in lighting and exposure by normalizing the brightness of compared images. This method is particularly effective in discerning subtle appearance similarities unaffected by environmental conditions.

$$\phi_0(x_a^i, x_a^j)(t) = \frac{\phi(x_a^i, x_a^j)(t)}{\sqrt{\phi(x_a^i, x_a^i)(0)\phi(x_a^j, x_a^j)(0)}}$$

**3. Sum of Absolute Differences (SAD):** By calculating the sum of absolute differences between feature vectors, SAD offers a straightforward yet powerful means of quantifying appearance disparities, proving invaluable in distinguishing between actors based on visual characteristics.

$$SAD(x_a^i, x_a^j) = \sum_k |x_a^{ik} - x_a^{jk}|$$

## 3.6.2. Positional Relation and Structural Insights

Positional information is crucial for understanding the spatial dynamics within a scene. We incorporate a distance mask to prioritize local interactions, employing a Euclidean distance threshold, μ, to filter relations based on proximity. This approach ensures that only actors within a meaningful distance influence each other's relational representation, reflecting the spatial structure of group activities.

Positional relations are captured using a distance mask to focus on locally significant interactions:

$$fs(x_s^i, x_s^j) = I(d(x_s^i, x_s^j) \leq \mu)$$

Where, I is the indicator function, d computes the Euclidean distance between actor bounding box centers, and μ is a predefined threshold.

### 3.6.3. Algorithmic Strategy and Temporal Integration

Our model extends beyond static relational analysis by constructing multiple ARGs, each tailored to capture different aspects of actor interactions. This multi-graph approach, combined with a temporal modeling strategy that samples frames sparsely, enables a dynamic representation of group activities over time. By integrating these ARGs with temporal information, our model achieves a comprehensive understanding of both the immediate and evolving nature of group dynamics.

## 3.7 Reasoning and Training on Graphs

In our project, after constructing the Actor Relation Graphs (ARGs), we advance to the crucial phase of relational reasoning and training. This stage is key to recognizing individual actions and group activities within the complex dynamics of multi-actor scenes. Our methodology employs Graph Convolutional Networks (GCN) for this purpose, leveraging their ability to perform computations directly on graph structures and efficiently aggregate relational information.

### 3.7.1 Graph Convolutional Network (GCN) Framework

**GCN Operation:** At the heart of our reasoning module is the GCN, which facilitates the feature aggregation across nodes (actors) by considering the graph's topology. A single layer of GCN can be mathematically expressed as:

$$Z^{(l+1)} = \sigma(GZ^{(l)}W^{(l)})$$

Here, $G \in R^{N\times N}$ represents the adjacency matrix of the ARG, reflecting the pairwise relations among actors. $Z^{(l)} \in R^{N\times d}$ denotes the feature matrix of the nodes at layer l, with $Z^{(0)} = X$ being the initial node features. $W^{(l)} \in R^{d\times d}$ is the weight matrix for layer l, and σ represents a non-linear activation function, such as ReLU, facilitating the layer-wise propagation of features.

**Graph Fusion Strategies**: Given the construction of multiple ARGs to encapsulate different relation aspects, we explore various fusion techniques post-GCN processing. Our approach primarily focuses on late fusion, where features of the same actor across different graphs are aggregated after individual GCN processing:

$$Z^{(l+1)} = \sum_{i=1}^{N_g} \sigma(G_i Z^{(l)} W^{(l,i)})$$

This element-wise sum ensures a comprehensive feature representation, integrating diverse relational insights. We also assess the potential of concatenation and early fusion, aiming to identify the most effective strategy for enhancing relational reasoning.

### 3.7.2 Training and Scene Representation

**Integration of Features:** Post-GCN, the aggregated relational features are combined with the original actor features to form a robust scene representation. This composite feature set serves as input to classifiers dedicated to predicting individual actions and the overall group activity, ensuring a holistic analysis of the scene dynamics.

**End-to-End Training:** Our model benefits from end-to-end training, utilizing backpropagation to optimize both the GCN parameters and the classifiers in a unified framework. The loss function is a combination of cross-entropy losses for both individual action classification (L1) and group activity recognition (L2), balanced by a weighting parameter $\boldsymbol{\lambda}$:

$$L = L_1(y_G, \hat{y}_G) + \lambda L_2(y_I, \hat{y}_I)$$

where ;
$y_G$ and $y_I$ are the ground-truth labels for group activity and individual actions, respectively, and $\widehat{yG}$ and $\widehat{yI}$ represent the corresponding predictions.

# CHAPTER 4: RESULTS

## 4.1 Introduction

In our research, we developed a comprehensive pipeline for "Computer Vision-Based Group Activity Detection and Action Spotting," focusing on the utilisation of backbone networks such as Inception V3, Mobile Net, and VGG16, with a particular emphasis on Normalised Cross-Correlation (NCC) for appearance similarity search. We expanded our methodology by varying the appearance similarity techniques—NCC, Sum of Absolute Differences (SAD), and dot product—specifically with Inception V3 as the backbone. Our evaluation was conducted using the Collective Activity dataset, which contains sequences of images, allowing us to assess the performance in both high and low crowded scenarios. Key metrics for our tests included accuracy, execution time, and resource consumption (RAM and GPU), providing a holistic view of the system's efficiency and effectiveness in real-world conditions.

## 4.2 Without Mask Integration

### 4.2.1 Trade-off Between Different Backbone networks

In the domain of computer vision, specifically for group activity detection, the choice of the backbone network for feature extraction is pivotal for achieving a balance between accuracy and computational efficiency. This analysis compares three widely used networks: Inception V3, MobileNet, and VGG16 (note: VGG19 results were provided, but we'll discuss VGG16 as initially mentioned, assuming a typo in the results section), focusing on their performance in both crowded and uncrowded scenarios. The comparison is based on accuracy and inference time per image is given below in Figure 14 and Figure 15.

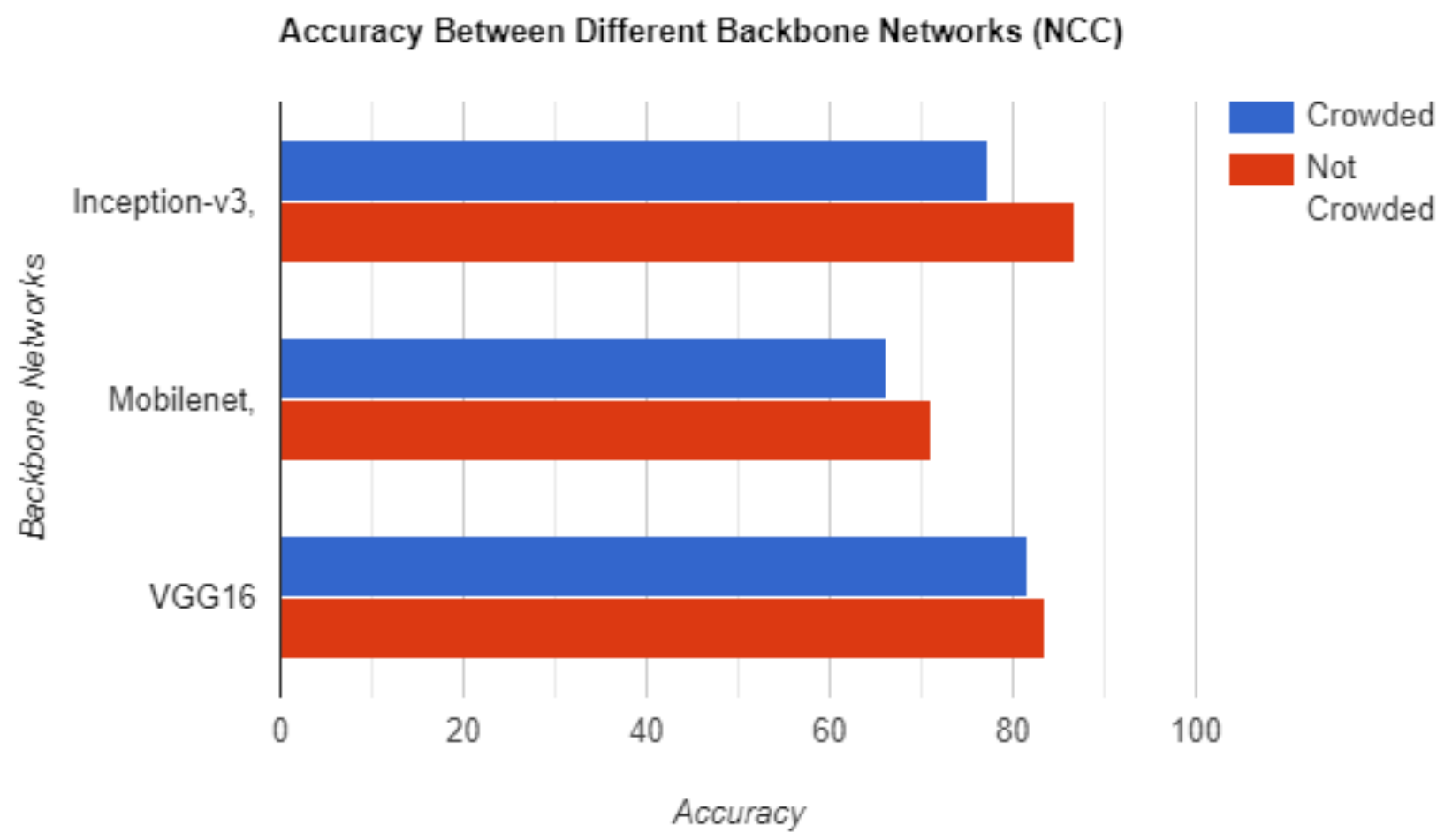

*Figure 4. 1: Accuracy Between different Backbones*

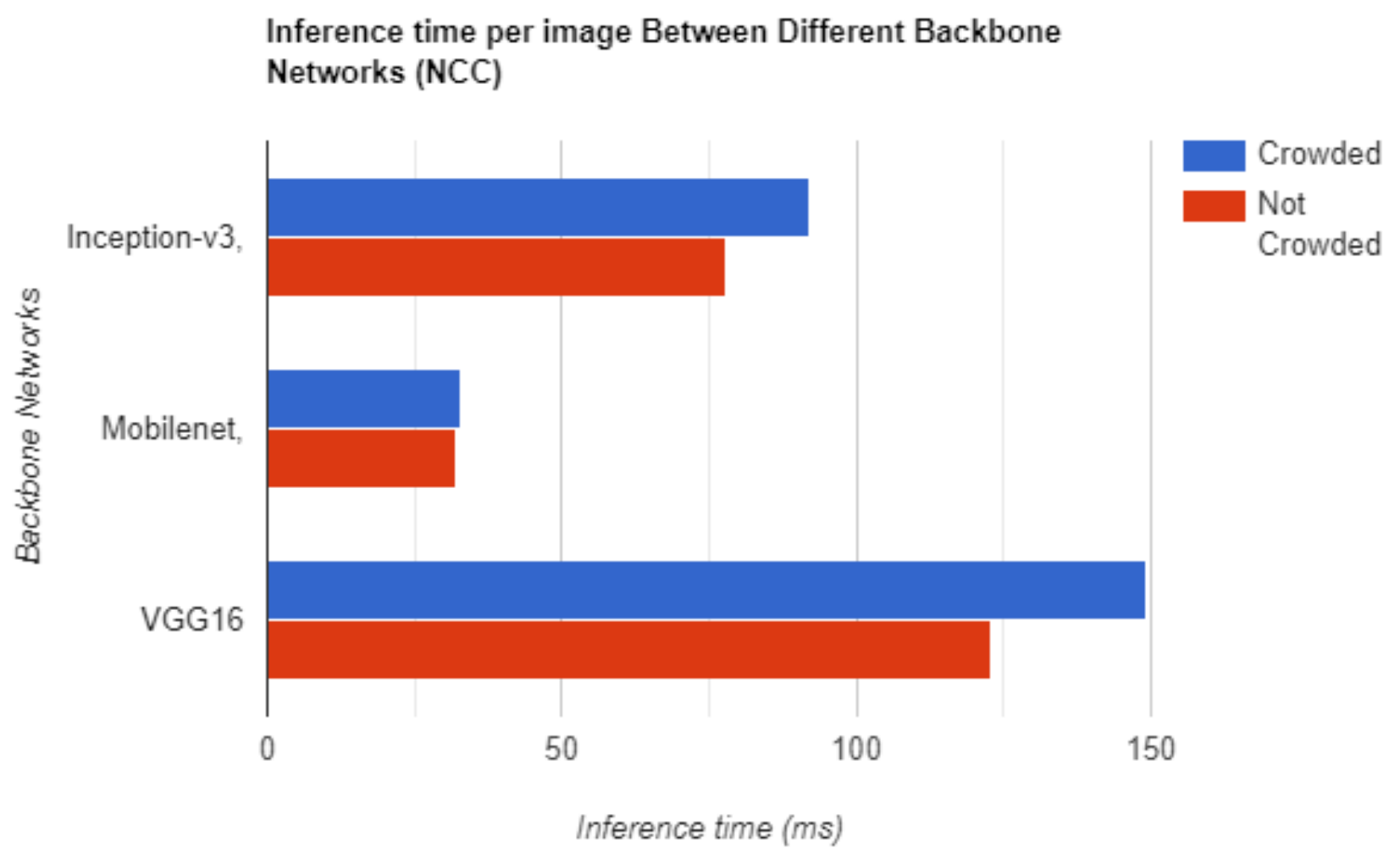

*Figure 4. 2: Inference time per image for different backbone networks (NCC)*

| **Accuracy, Inference time per image** | **Inception v3** | **Mobile net** | **VGG19** |
|---|---|---|---|
| **Not crowded scenario** | 86.72%<br>78ms | 71.02%<br>32ms | 83.64%<br>123ms |
| **crowded scenario** | 77.43%<br>92ms | 66.13%<br>33ms | 81.666%<br>149ms |

*Table 4. 1: Accuracy and Execution time for different backbone networks using NCC as similarity search*

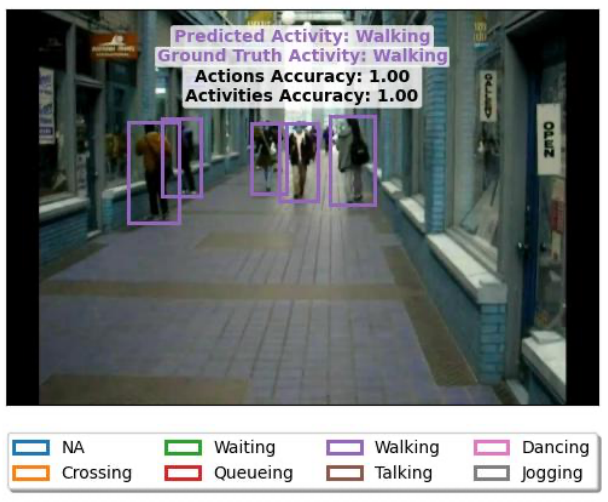

Inception-v3

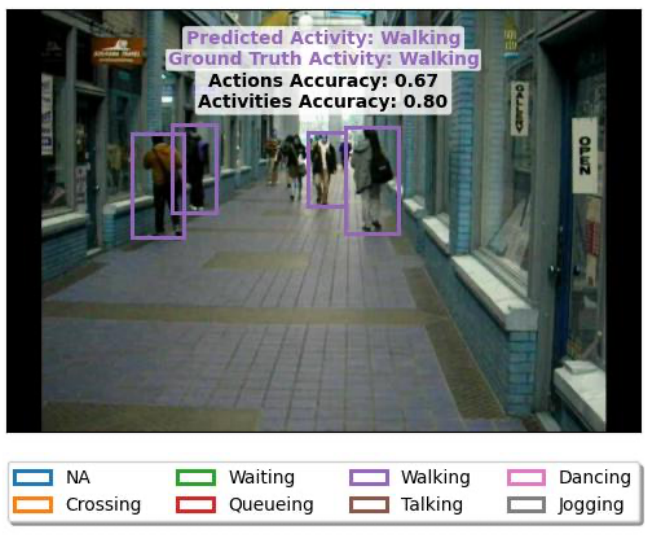

Mobilenet

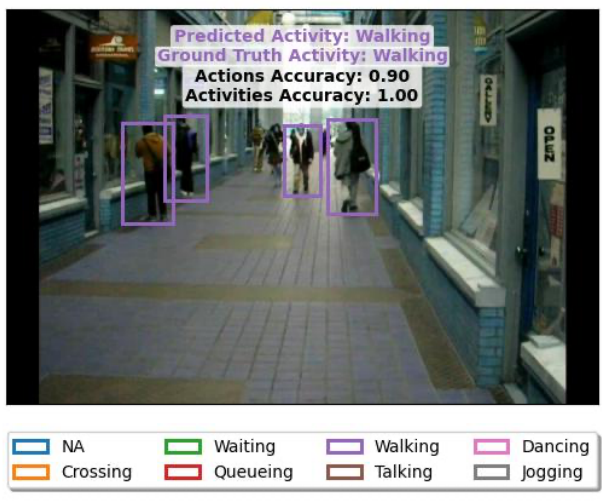

VGG16

*Figure 4. 3 Outputs for different Backbone Networks for Crowded scenes*

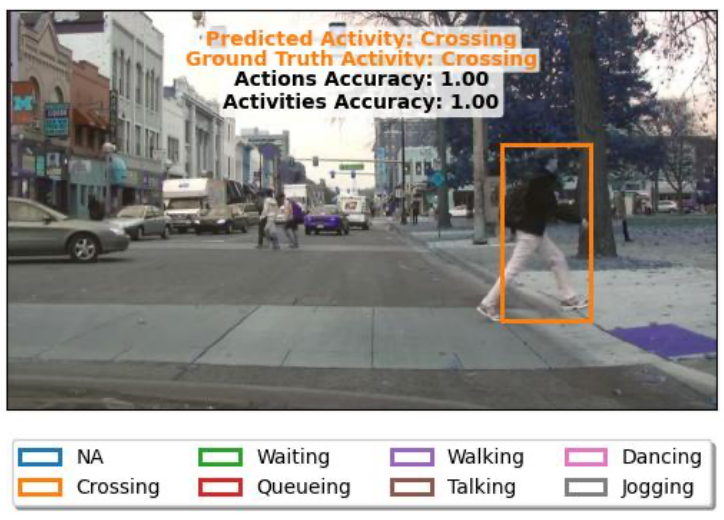

Inception-v3

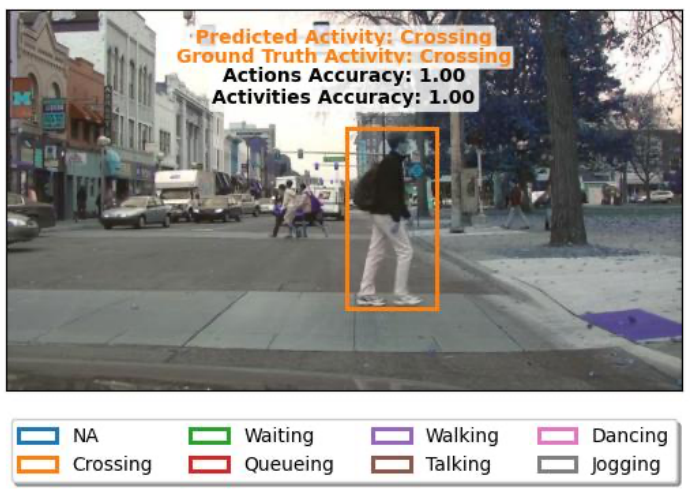

Mobilenet

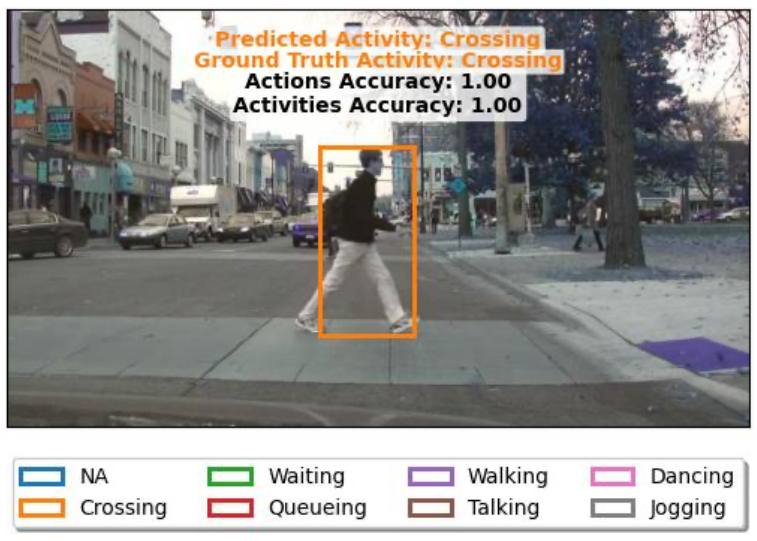

VGG16

*Figure 4. 4: Outputs for different Backbone Networks for Not Crowded scenes*

Inception V3 is known for its depth and complexity, incorporating multiple kernel sizes within the same layer to capture information at various scales. This architecture leads to high accuracy due to its ability to learn more complex features, but at the cost of longer inference times. In non-crowded scenarios, Inception V3 achieves an accuracy of 86.72% with an inference time of 78ms. In crowded scenarios, the accuracy drops to 77.43%, with an inference time of 92ms. The drop in accuracy in crowded scenes is expected due to the increased complexity of the scene, but Inception V3 handles it relatively well, maintaining reasonable accuracy.

MobileNet, designed for mobile and embedded vision applications, focuses on efficiency. It uses depth-wise separable convolutions to reduce the number of parameters and computational cost, trading off a bit of accuracy for speed. This trade-off is evident in the results, where MobileNet achieves 71.02% accuracy in non-crowded scenarios with a fast inference time of 32ms, and 66.13% accuracy in crowded scenarios with a similarly low inference time of 33ms. MobileNet's strength lies in its efficiency, making it suitable for applications where speed is critical, and absolute accuracy can be slightly compromised.

VGG16 (assuming the results for VGG19 were meant for VGG16), a simpler and older architecture characterized by its uniform use of 3x3 convolutional layers stacked deeply, shows a different trade-off. It's less efficient compared to MobileNet, with longer inference times (123ms in non-crowded and 149ms in crowded scenarios) but achieves higher accuracy than MobileNet, with 83.64% in non-crowded and 81.666% in crowded scenarios. VGG16's architecture, while straightforward, is computationally intensive due to the number of parameters, leading to longer inference times but also providing robust feature extraction capabilities.

Comparing these results that is shown in Figure 16 and Figure 17 with state-of-the-art (SOTA) models in object detection for group activity detection, we observe that each chosen backbone has its niche. Inception V3 offers a balanced approach with high accuracy and moderate speed, suitable for applications where detailed feature extraction is critical, and some computational delay is acceptable. MobileNet excels in scenarios where rapid inference is crucial, sacrificing some accuracy for speed. VGG16 stands as a middle ground, offering high accuracy at the cost of increased computational requirements.

To conclude, the choice of backbone network for feature extraction in group activity detection should be guided by the specific requirements of the application, such as the need for high accuracy or fast inference times. While newer architectures may offer improvements in both dimensions, the presented results provide a solid baseline for understanding the trade-offs involved in selecting a backbone network for this application.

## 4.2.2 Trade-off Between different appearance similarity searches

In the realm of computer vision, especially for group activity detection, the method of appearance similarity search is fundamental for effective feature extraction and object detection. Three common techniques include Normalized Cross Correlation (NCC), Sum of Absolute Differences (SAD), and Dot Product. Each method offers distinct trade-offs in terms of accuracy and computational efficiency, impacting their suitability for various applications. Figure 18 describes the comparison of these methods based on research findings, focusing on their performance in both crowded and non-crowded scenarios.

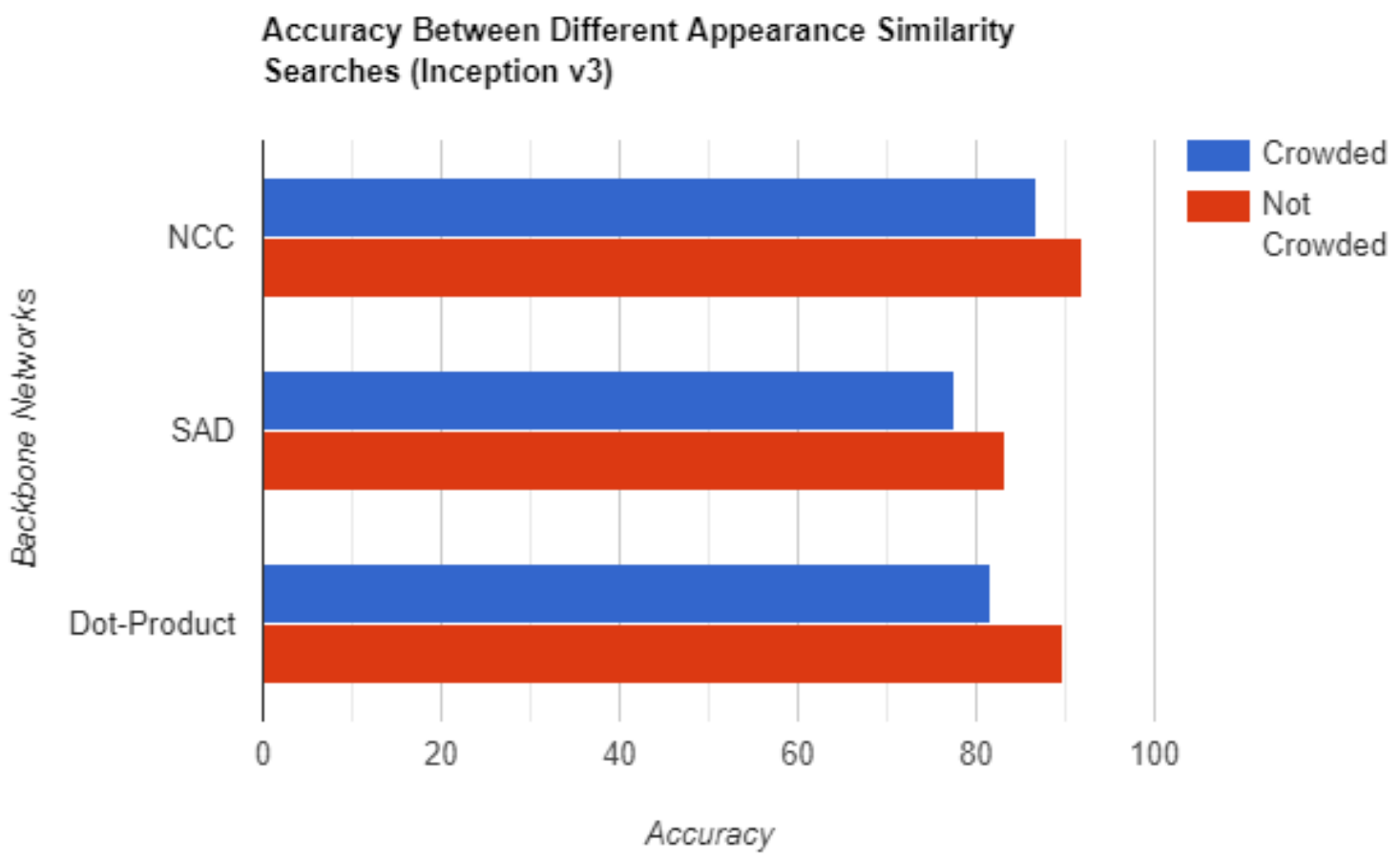

*Figure 4. 5: Accuracy between different appearance similarity searches*

| | NCC (Normalized Cross Correlation) | SAD (Sum Absolute Difference) | Dot Product |
|---|---|---|---|
| **Accuracy for crowded scenario** | 86.79% | 77.52% | 81.62% |
| **Accuracy for not crowded scenario** | 91.93% | 83.17% | 89.62% |

*Table 4. 2: Accuracy for different similarity searches using inception v3 as backbone network*

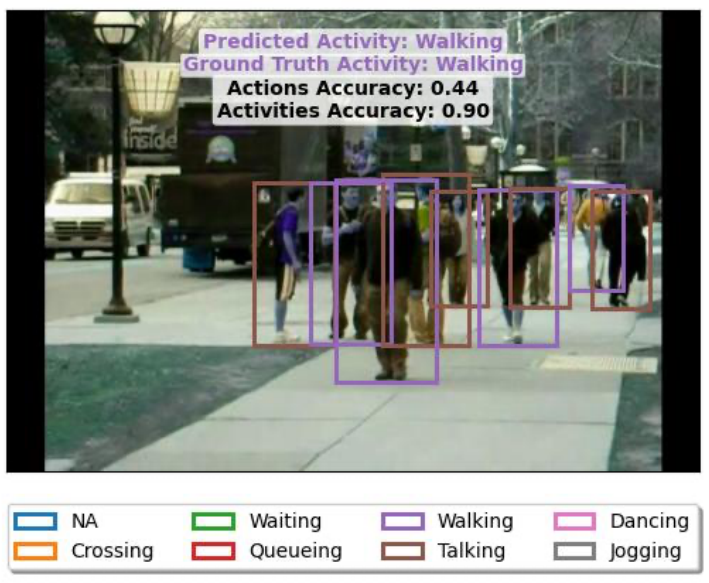

Dot Product

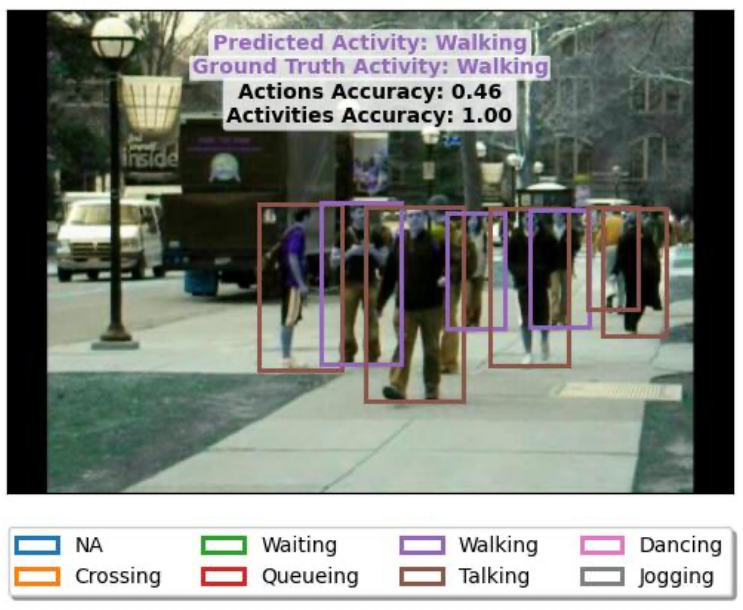

NCC

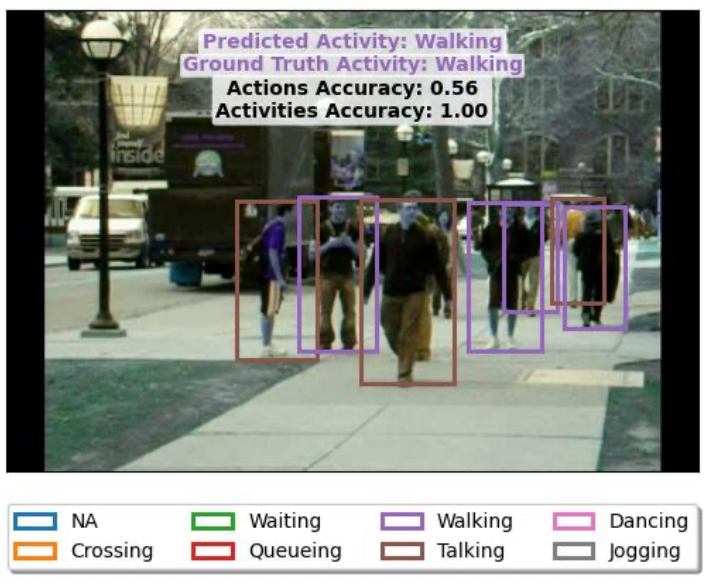

SAD

*Figure 4. 6: Outputs for different Similarity searches for Crowded scenes*

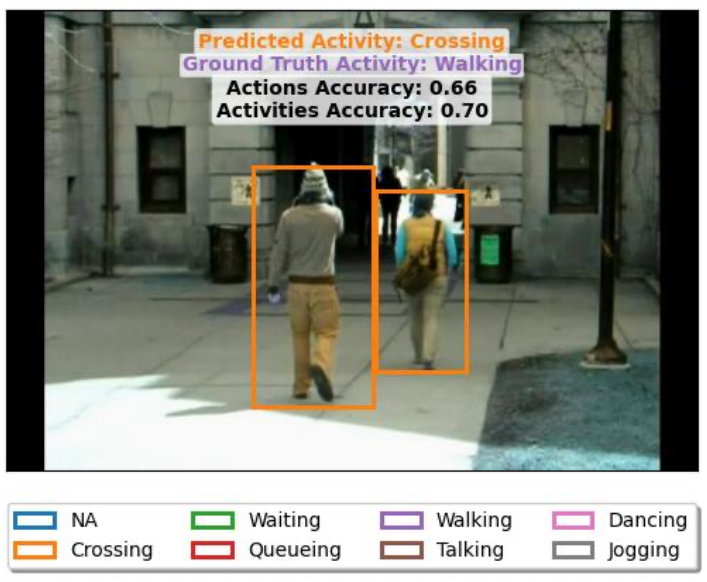

Dot Product

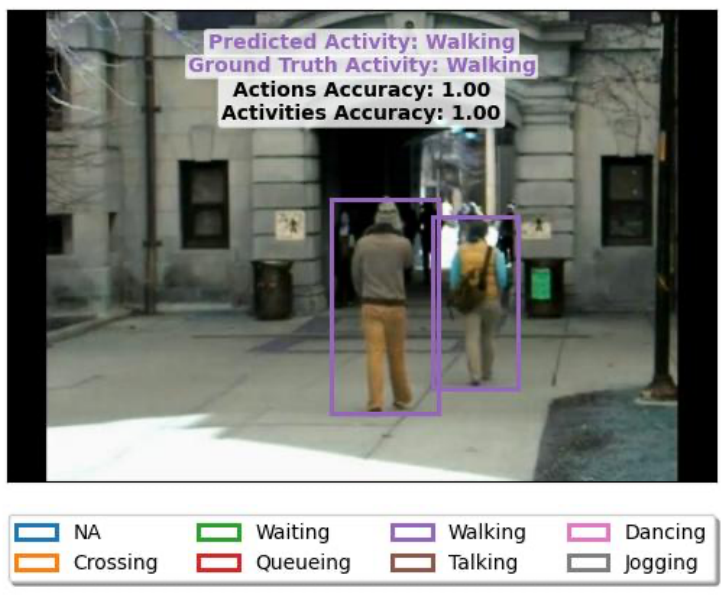

NCC

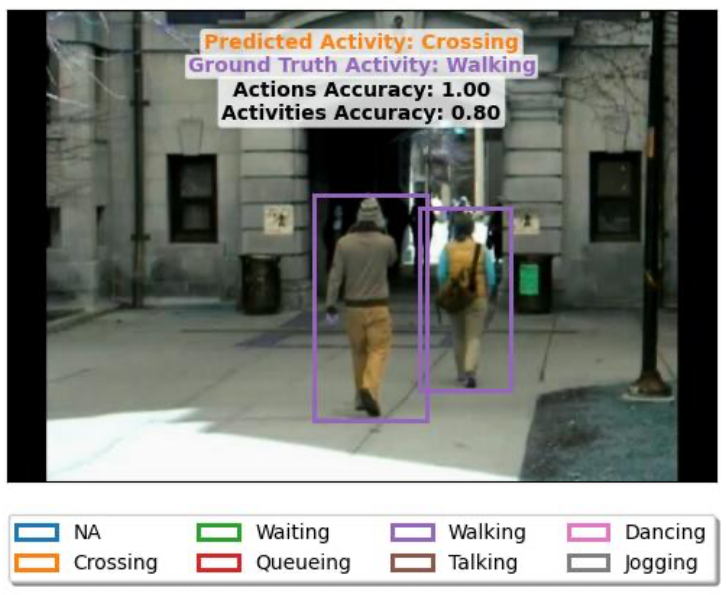

SAD

*Figure 4. 7: Outputs for different Similarity searches for Not Crowded scenes*

Normalized Cross Correlation (NCC) is a measure of similarity between two images that is robust to changes in lighting and exposure, making it particularly useful for matching features under varying conditions. NCC has demonstrated high accuracy in both crowded (86.79%) and non-crowded (91.93%) scenarios. The strength of NCC lies in its ability to provide a normalized metric for comparison, which is less sensitive to the absolute differences in intensity between the images being compared. This robustness makes NCC highly accurate, although it can be computationally more intensive than simpler methods due to the normalization process.

Sum of Absolute Differences (SAD) involves calculating the absolute difference between each pair of corresponding pixels in two images or image regions, then summing those differences. This method is straightforward and fast, making it appealing for real-time applications. However, the simplicity of SAD comes at the cost of accuracy, particularly in complex scenarios. The research results reflect this, with SAD achieving lower accuracy than NCC in both crowded (77.52%) and non-crowded (83.17%) scenarios. SAD's performance dip in crowded scenes can be attributed to its sensitivity to noise and variations in lighting, which are more prevalent in such conditions.

Dot Product measures the similarity between two vectors (in this case, image pixel intensity vectors) and is a common operation in many computer vision algorithms, particularly those involving neural networks. The dot product approach yielded an accuracy of 81.62% in crowded scenarios and 89.62% in non-crowded scenarios. This method's effectiveness hinges on the alignment and magnitude of the vectors, offering a balance between computational efficiency and accuracy. While not as robust to variations as NCC, the dot product can be computed more rapidly, making it suitable for applications where speed is a consideration, but some environmental control is possible to mitigate its sensitivity issues.

When comparing these results, shown in Figure 19 and Figure 20, with the state-of-the-art (SOTA) methods in object detection for group activity detection, it's evident that each technique has its niche. NCC stands out for its high accuracy, especially beneficial in applications where precise matching under varied conditions is crucial. SAD offers a speed advantage, ideal for real-time or embedded systems where computational resources are limited. Dot Product finds a middle ground, offering a compromise between speed and sensitivity to conditions, suitable for scenarios where moderate control over environmental factors is possible.

## 4.3 Comparison between feature map filtered through boundary boxes and masks

**Early Layers (Mixed0 Layer) -** The Mixed0 layer, shown in Figure 21 represents the first inception block in the Inception V3 model, is responsible for extracting low-level features from the input image. These feature maps primarily capture basic patterns such as edges, corners, and textures. The filters within this layer are tuned to detect simple structures, providing the model with a foundational understanding of the image content. The output from this layer consists of multiple feature maps, each highlighting a different aspect of the image's low-level features.

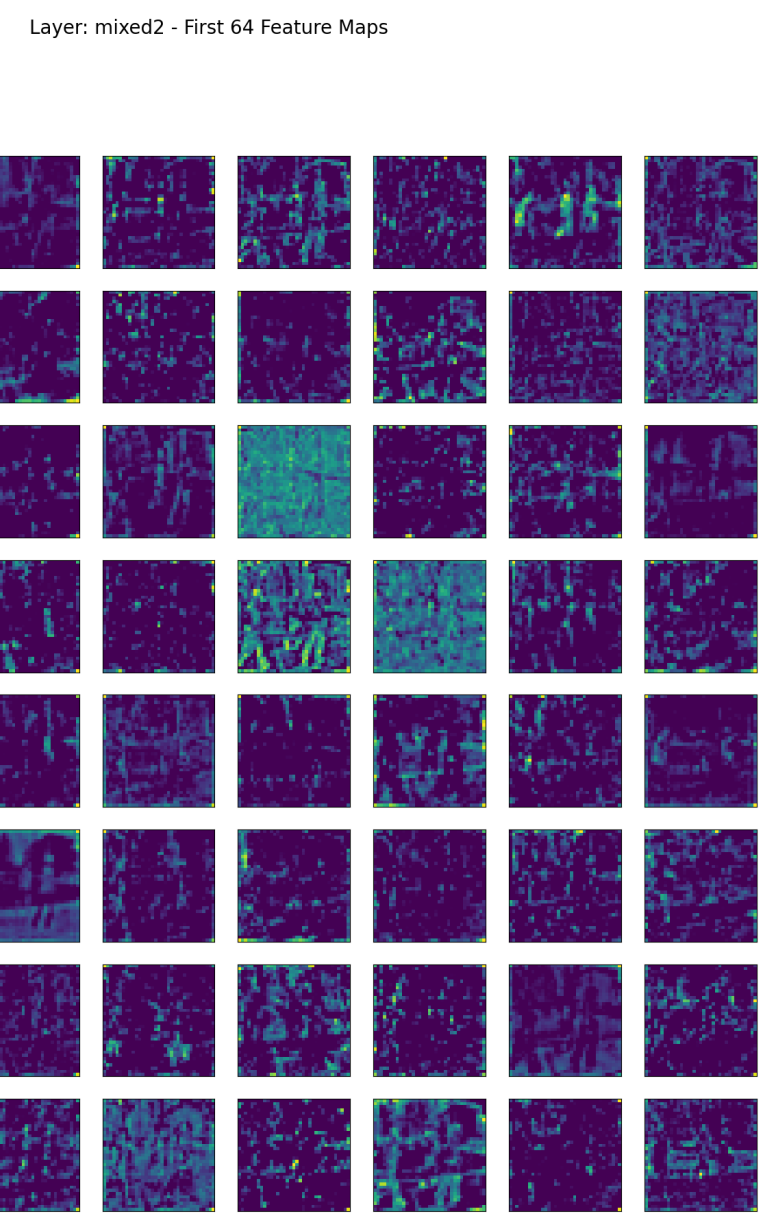

*Figure 4. 8: Feature maps of Early Layers (Mixed0 Layer)*

**Late Layers (Mixed10 Layer) -** In contrast, the Mixed10 layer, shown in Figure 22, represents the final inception block and is responsible for extracting high-level features from the input image. These feature maps capture complex and abstract representations that are essential for distinguishing between different objects and scenes within the image. The filters in this layer can identify intricate patterns and advanced features that go beyond basic image structures, making them crucial for the model's decision-making process in identifying group activities and spotting actions.

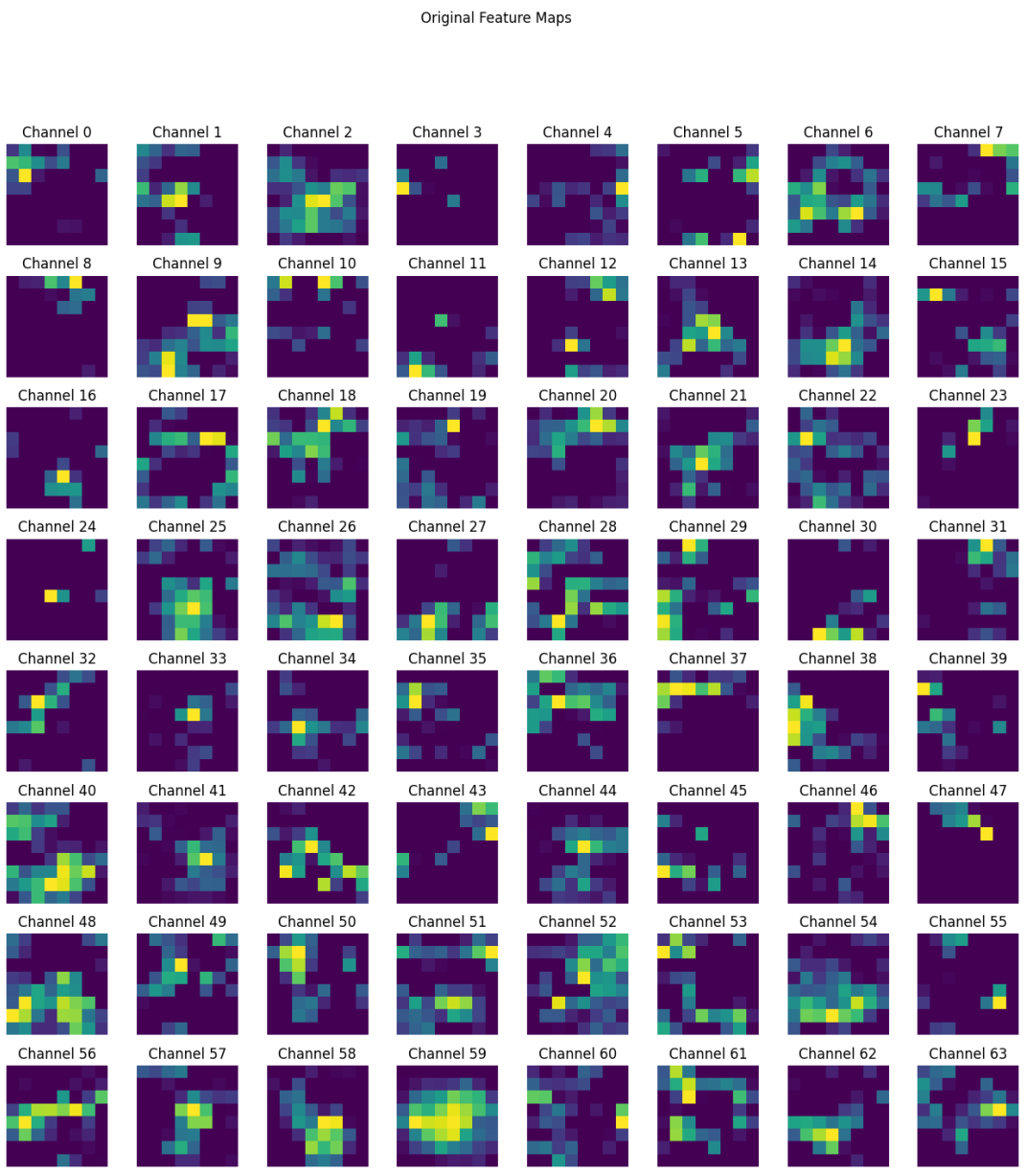

*Figure 4. 9: Feature maps of Late Layers (Mixed10 Layer)*

In our project, we aimed to enhance the accuracy of group activity detection and action spotting by leveraging advanced techniques in feature extraction and filtering. Utilizing the Inception V3 model, we focused on capturing and analyzing feature maps from both early and late stages of the network. To improve the precision of our analysis, we implemented a mask filtering approach that allows us to isolate relevant information within specific regions of interest (ROIs). This method enables us to refine the features corresponding to individual persons within the image, ensuring that our system accurately identifies and interprets group activities.

This approach involves a detailed process of extracting feature maps from crucial layers within the Inception V3 model. We concentrated on two key layers: the first inception block (Mixed0 layer) and the final inception block (Mixed10 layer). These layers were selected to capture both low-level features, such as edges and textures, as well as high-level, abstract features that are essential for distinguishing between different objects and activities in the image. After extracting these feature maps, we applied mask filters to isolate and highlight pertinent information, further refining the analysis of individual persons within the scene.

To enhance the focus on specific regions of interest within the image, we applied mask filters to the extracted feature maps. These masks are meticulously designed to align with the regions corresponding to individual persons, ensuring that irrelevant information is effectively filtered out. The process begins with identifying the ROIs within the feature maps and then applying the masks to cover these areas precisely. This targeted approach allows us to concentrate on the most critical parts of the image, improving the overall accuracy of the analysis.
Following the application of masks, we proceed to isolate the features that are most relevant to the identified ROIs. By filtering out extraneous details, this step ensures that only the essential information is retained for further processing. The isolated features correspond directly to the regions where individual persons are located, which is crucial for accurate group activity detection and action spotting. This focused isolation of features not only enhances the precision of our system but also reduces computational overhead by eliminating unnecessary data from the analysis.

In our ongoing efforts to enhance object detection and activity recognition in computer vision, we conducted a comparative analysis of two feature filtering techniques: bounding box filtering and mask filtering. Utilizing the Inception V3 model, we focused on extracting feature maps at critical stages of the network, particularly the first inception block (Mixed0 layer) and the final inception block (Mixed10 layer). By applying both bounding box and mask filters to isolate relevant information for individual persons within images, this study aims to determine which method yields more precise and accurate features. The findings from this analysis are intended to inform future improvements in our object detection capabilities.

Our analysis involved a systematic approach, beginning with the extraction of feature maps from the Inception V3 model. We then applied both bounding box and mask filtering techniques to these feature maps, isolating relevant features corresponding to individual persons within the images. The final step involved a detailed comparison of the filtered features to assess the precision and accuracy of each method.

## Comparison and Analysis

- **Filtered Feature Maps (Mixed0 Layer)**

Bounding Box Filtering - The bounding box filtered feature maps from the first inception block (Mixed0 layer) demonstrate how the model identifies basic structures and shapes specific to each person. This method provides a general outline of the relevant areas, capturing fundamental features such as edges and textures.

Mask Filtering - In contrast, the mask filtered feature maps from the Mixed0 layer exhibit a more refined isolation of these basic structures. By reducing noise and excluding irrelevant information, the mask filtering technique offers a clearer and more focused representation of the low-level features within the image.

- **Filtered Feature Maps (Mixed10 Layer)**

Bounding Box Filtering - The bounding box filtered feature maps from the final inception block (Mixed10 layer) reveal high-level features that highlight complex patterns and objects specific to each person. This method captures detailed and abstract representations, essential for distinguishing between different objects and activities.

Mask Filtering - The mask filtered feature maps from the Mixed10 layer provide even more precise and accurate representations. By focusing exclusively on the most relevant regions within the ROIs, mask filtering delivers finer detail and superior feature isolation, enhancing the model's ability to detect and interpret complex features.

Our analysis clearly demonstrates that the mask filter provides superior accuracy compared to the bounding box filter. By precisely targeting and isolating the most relevant regions within the feature maps, the mask filter effectively reduces noise and eliminates irrelevant information. This leads to a more refined and accurate representation of the features, particularly in complex and high-level scenarios

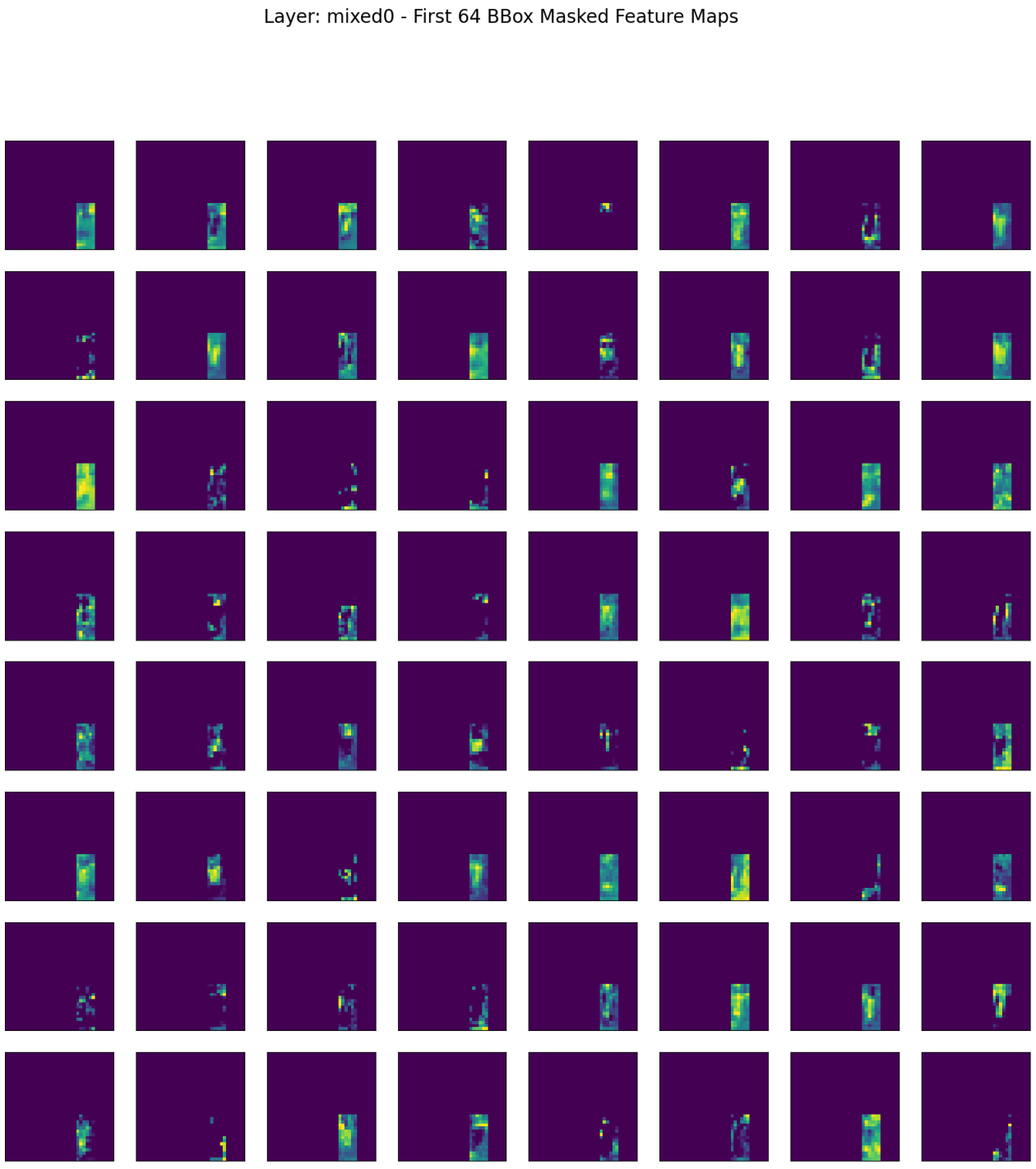

*Figure 4. 10: Feature maps of Bounding Box Filtering (Mixed0 Layer)*

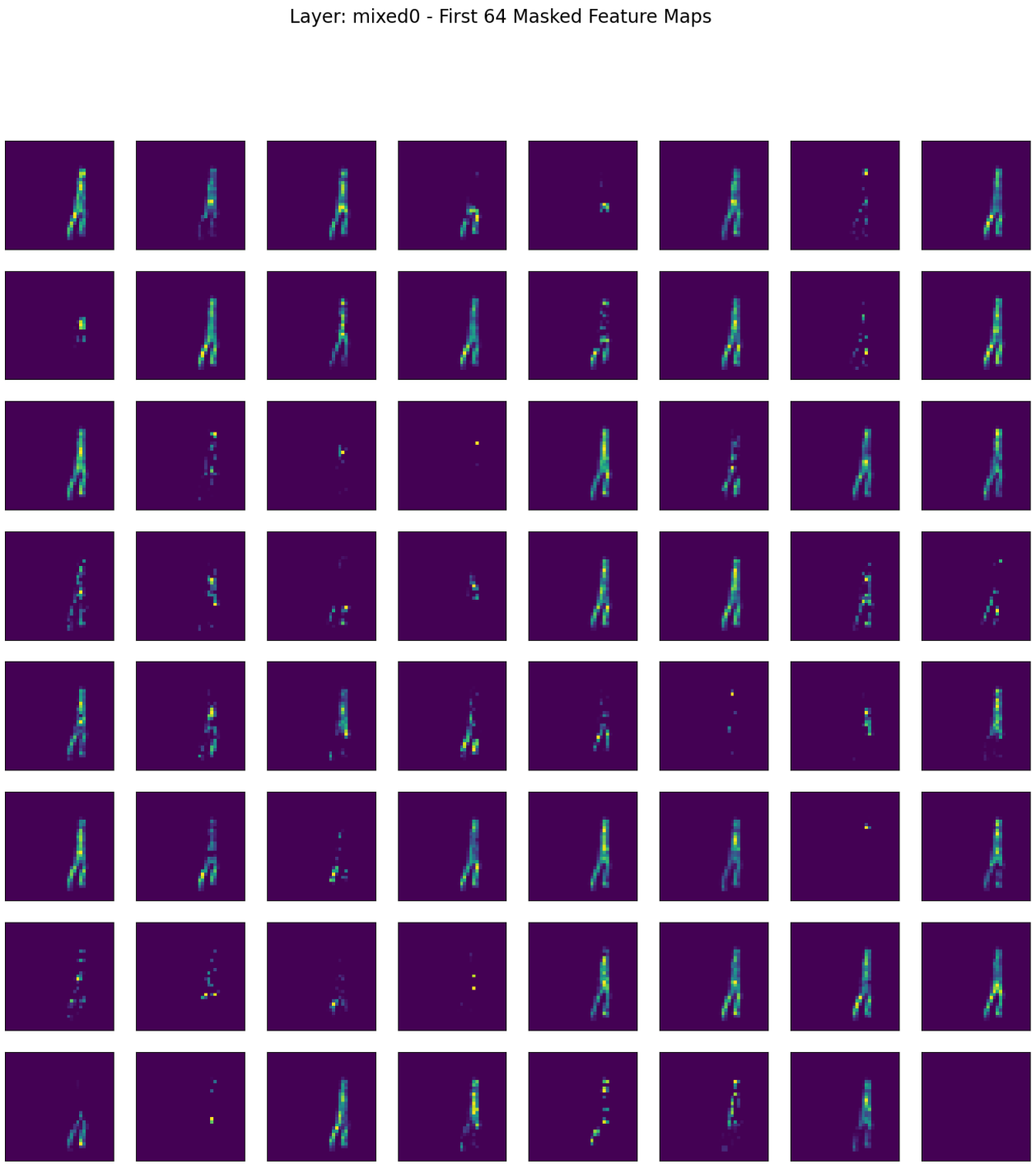

*Figure 4. 11: Feature maps of mask Filtering (Mixed0 Layer)*

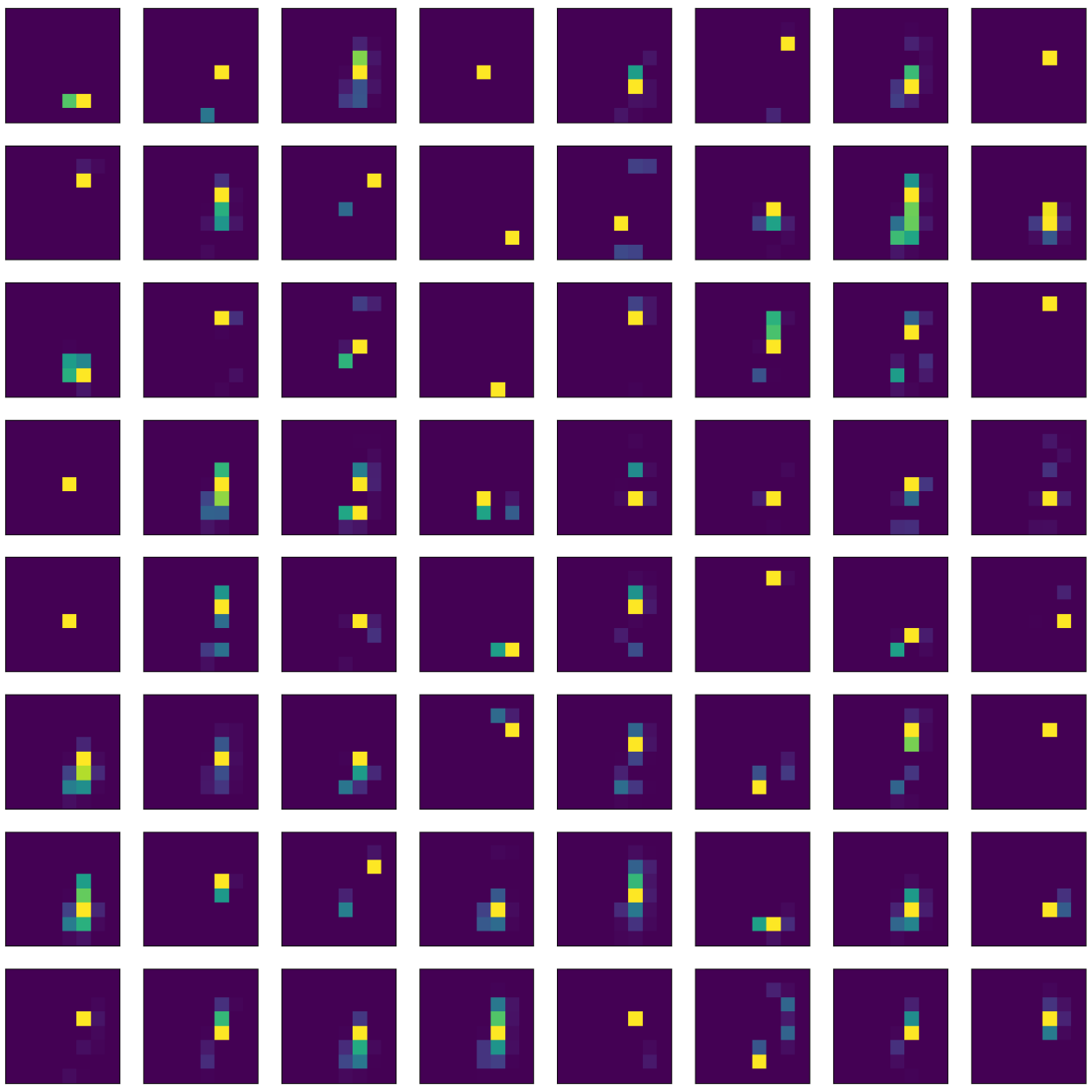

*Figure 4. 12: Feature maps of Bounding Box Filtering (Mixed10 Layer)*

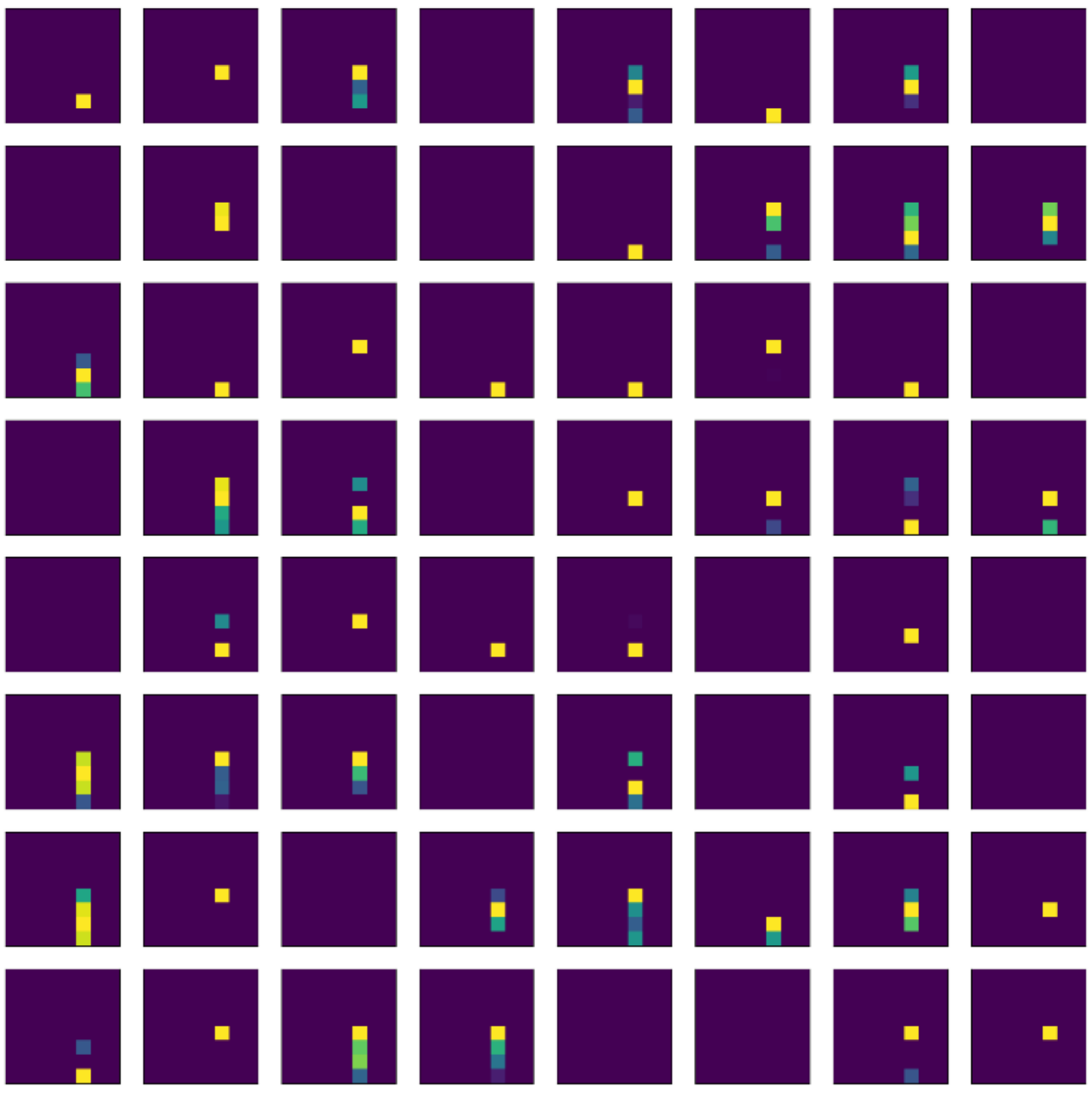

*Figure 4. 13: Feature maps of Mask Filtering (Mixed10 Layer)*

The above Figure 25 shows the features filtered through the bounding box, while the other Figure 26 shows the features filtered through masking. It is clear that unwanted features are effectively filtered out using the mask filter.

## Comparison between final output and accuracy between boundary boxes and mask filters

**Figure 25 illustrates the output from bounding Box filter.**

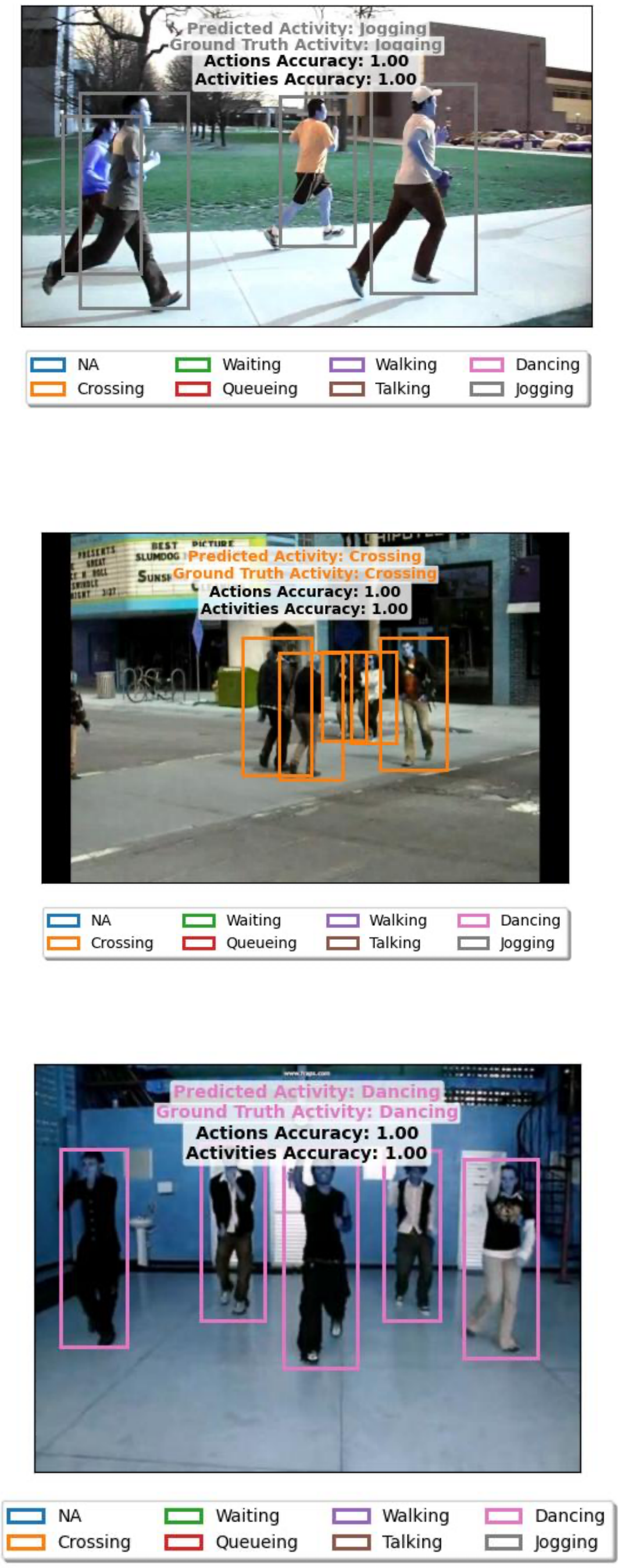

*Figure 4. 14: Output from Bounding Box filter*

```
====> Test at epoch #20
Group Activity Accuracy: 81.62%, Individual Actions Accuracy: 78.47%
```

**Figure 26 illustrates the output from mask filter.**

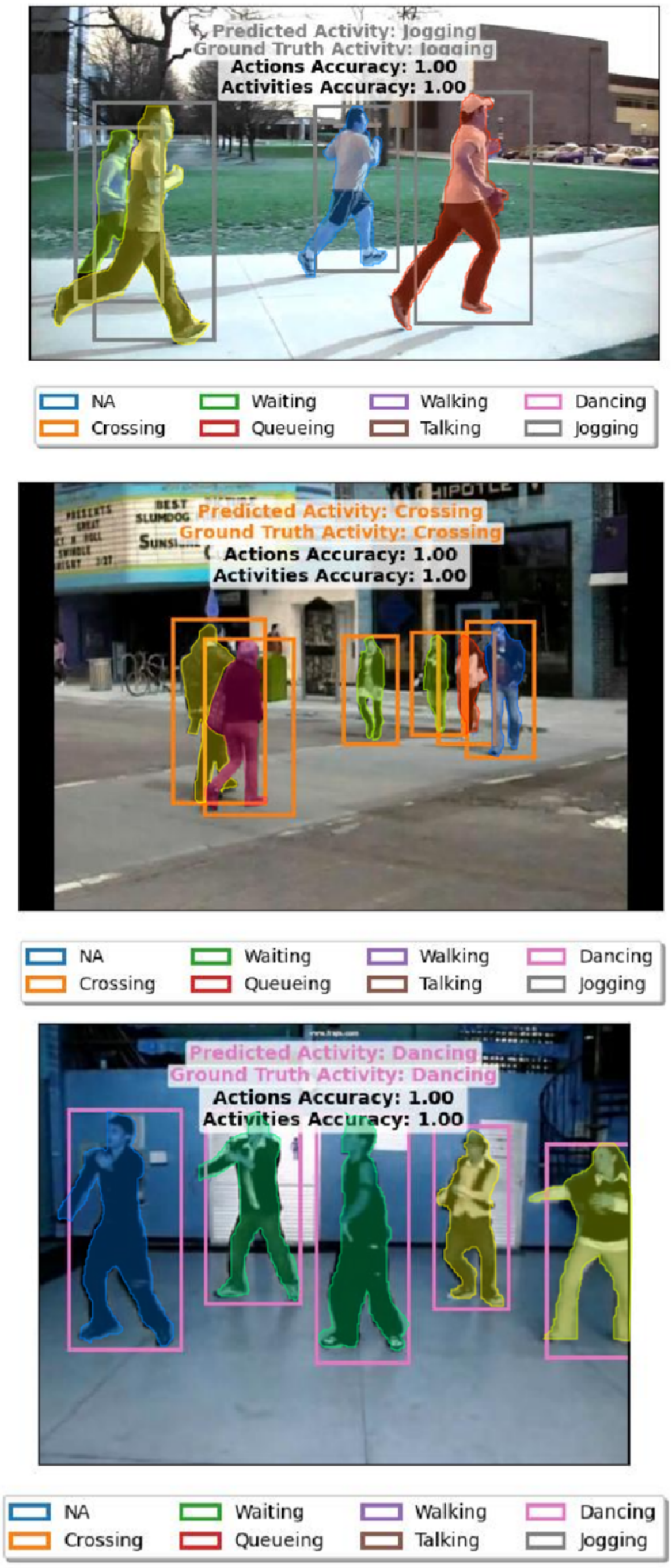

*Figure 4. 15: Output from Mask filter*

```
====> Test at epoch #20
Group Activity Accuracy: 84.15%, Individual Actions Accuracy: 80.89%, Loss: 1.75282
```

From the above results the accuracy through bounding box filter gives an accuracy of 81.62% and accuracy through mask filter is 84.15%. so, mask filter gives better accuracy when compared to features extracted through bounding box filter.

# CHAPTER 5: CONCLUSION

Based on the detailed examination of our improved final year project on group activity recognition, which incorporated VGG16 alongside MobileNet and Inception V3 for backbone networks, and utilised Normalised Cross-Correlation (NCC) in addition to Sum of Absolute Differences (SAD) and dot product for appearance similarity calculation, and mask filtering for feature refinement, here's a structured conclusion:

- **Comparative Analysis of Backbone Networks:** Our experiments demonstrated that each backbone network contributes uniquely to the model's performance. VGG16, alongside the lightweight MobileNet and the efficient Inception V3, provided a comprehensive understanding of different architectures' impacts on accuracy and computational efficiency.

- **Advanced Appearance Similarity Calculation:** Employing NCC alongside SAD and dot product methods for appearance similarity calculation introduced a multifaceted perspective to understanding actor relations. This diversity in methods enriched our model's ability to accurately recognize group activities.

- **Mask Filtering for Feature Refinement:** The integration of mask filtering significantly enhanced the precision of feature extraction, reducing noise and improving the overall accuracy of the model, particularly in complex scenarios where accurate feature isolation is critical.

- **Performance Evaluation:** The combination of advanced backbone networks and multifaceted appearance similarity calculations significantly enhanced the model's performance, showcasing the effectiveness of the proposed improvements.

- **Contribution to the Field:** This project represents a significant step forward in video understanding and group activity recognition, advancing the current state-of-the-art and opening new pathways for future research.

In conclusion, our final year project successfully demonstrates the effectiveness of incorporating diverse backbone networks and appearance similarity calculations in improving group activity recognition, contributing valuable insights into the field and laying the groundwork for future innovations in video understanding technologies.

# CHAPTER 6: REFERENCES